\documentclass{article}

\newcommand*\DatasetsTable{
    \begin{table}
        \centering
        \tiny
        \caption{Details of different datasets used in our benchmark }
        \begin{tabular}{lcccll}
        \toprule
             \textbf{Dataset} & \multicolumn{1}{l}{\textbf{\#Samples}} & \multicolumn{1}{l}{\textbf{\#Class}} & \multicolumn{1}{l}{\textbf{Patch Size}} &  \textbf{Tissue Types} & \textbf{Other Information} \\ 
             
        \midrule 
        \begin{tabular}[c]{@{}l@{}}\textbf{\href{https://warwick.ac.uk/fac/sci/dcs/research/tia/data/crc-tp}{CRC-TP \cite{javed2020multiplex}}}\\(Colon Cancer) \\ (\href{https://warwick.ac.uk/fac/cross_fac/tia/data/crc-tp}{Ready to use})\end{tabular} & 280,000 & 7 &  $150\times150$ & \begin{tabular}[c]{@{}l@{}} Tumor, Stroma, Complex Stroma \\ Muscle, Debris, Inflammatory, \\ Benign \end{tabular} & \begin{tabular}[c]{@{}l@{}} - Train: 70\%  - Test: 30\% \\  -- 91\% F1 Score \cite{javed2020cellular} \\ - Train: 14 - Test: 6 (patients) \\ -- 84\% F1 Score \cite{javed2020cellular} \end{tabular} \\

        \midrule
        \begin{tabular}[c]{@{}l@{}}\textbf{\href{https://doi.org/10.5281/zenodo.1214456}{NCT \cite{kather100_data}}} \\ (Colon cancer)\\ (\href{https://doi.org/10.5281/zenodo.1214456}{Ready to use})\end{tabular} & 100,000 & 9 & $224\times224$ & \begin{tabular}[c]{@{}l@{}}Adipose, Background, Debris \\ Lymphocytes, Mucus, Smooth Muscle \\ Normal Colon Mucosa, \\ Cancer-associated Stroma, \\ Colorectal Adenocarcinoma Epithelium\end{tabular} & \begin{tabular}[c]{@{}l@{}}Two different splits: \\ - 70\% 15\% 15\% \\ -- 98.7\% Accuracy \cite{kather2019predicting}\\ - 7180 external test set\\ -- 94.3\% Accuracy \cite{kather2019predicting}\end{tabular} \\ 
        
        \midrule
        \begin{tabular}[c]{@{}l@{}}\textbf{\href{https://github.com/tampapath/lung_colon_image_set}{LC25000 \cite{borkowski2019lung}}} \\ (Colon and Lung cancer) \\  (\href{https://academictorrents.com/details/7a638ed187a6180fd6e464b3666a6ea0499af4af}{Ready to use})\end{tabular} & \begin{tabular}[c]{@{}c@{}}1'250\\(original)\\25'000\\ (augmented)\end{tabular} & 5 &  $768\times768$ & \begin{tabular}[c]{@{}l@{}} Benign lung, Benign colon,\\ Lung adenocarcinoma, \\ Squamous cell carcinomas, \\ Colon adenocarcinoma \end{tabular} &  \begin{tabular}[c]{@{}l@{}} - 75\% train 25\% test\\ -- 98.7\% Accuracy \cite{sarwinda2020analysis} \\ - 85\% train 15\% test\\ -- 98.5\% Accuracy \cite{sarwinda2020analysis}\end{tabular}   \\ 
        
        \midrule
        \begin{tabular}[c]{@{}l@{}}\textbf{\href{https://web.inf.ufpr.br/vri/databases/breast-cancer-histopathological-database-BreakHis/}{BreakHis \cite{spanhol2015dataset}}} \\ (Breast cancer)\\  (\href{https://web.inf.ufpr.br/vri/databases/breast-cancer-histopathological-database-BreakHis/}{Available upon request})\end{tabular} & 7,909 & \begin{tabular}[c]{@{}c@{}}8 \\ (4 Malignant\\ + 4 Benign)\end{tabular} & $700\times460$ & \begin{tabular}[c]{@{}l@{}}Adenosis, Fibroadenoma,  Phyllodes-\\Tumor, Tubular Adenona, Carcinoma\\Lobular Carcinoma, Mucinous- \\Carcinoma, Papillary Carcinoma \end{tabular} & \begin{tabular}[c]{@{}l@{}} Different magnifying factors: \\ - 40X, 100X, 200X, 400X \\ - 80\% train 20\% test \\ -- 98.8\% Accuracy \cite{tougaccar2020breastnet}  \end{tabular} \\ \hline 
        
        \end{tabular}
        \label{tab:datasets}
    \end{table}
}

\newcommand*\MainTable{
    \begin{table}
        \centering
        \small
        \caption{Test accuracy in three different scenarios; \neardomain (CRC-TP $\rightarrow$ NCT), \middledomain (CRC-TP $\rightarrow$ LC25000) and \outdomain (CRC-TP $\rightarrow$ BreakHis). The backbone network in all the experiments is ResNet-18. All results are averaged over three seeds. }
        \begin{tabular}{lccccc}
            \specialrule{1.5pt}{1pt}{1pt}
            & &  \multicolumn{3}{c}{\textbf{CRC-TP $\rightarrow$ NCT}} \\
             Method & Type &  1-shot & 5-shot & 10-shot\\
             \toprule
             ProtoNet \cite{prototypical_nets} & \multirow{5}{*}{Inductive} & 58.1 $\pm 0.69$ & 72.3 $\pm 0.53$ & 74.4 $\pm 0.51$ \\
             MetaOpt \cite{lee2019meta} & & 48.6 $\pm 0.71 $ & 64.4 $\pm 0.56 $ & 70.0 $\pm 0.52 $\\
             SimpleShot \cite{simpleshot} & & 63.2 $\pm 0.66 $ & 76.1 $\pm 0.52 $ & 78.4 $\pm 0.5 $\\
             Distill \cite{tian2020rethinking} &  &  61.9 $\pm 0.64$ & 77.3 $\pm 0.49$ & 80.9 $\pm 0.47$\\
             Finetune \cite{closer_look} & & \textbf{64.7 $\pm 0.65$} & \textbf{81.4 $\pm 0.48$} & \textbf{85.0 $\pm 0.42$}  \\
             \midrule
             MAML \cite{maml} & \multirow{3}{*}{Transductive} &  53.6 $\pm 0.76$ & 70.5 $\pm 0.55$ & 74.5 $\pm 0.51$\\
             LaplacianShot \cite{Laplacian} &  &  62.8 $\pm 0.65$ & 76.2 $\pm 0.51$ & 78.1 $\pm 0.5$\\
             TIM \cite{TIM} &  & \textbf{70.9 $\pm 0.85 $} & \textbf{83.5 $\pm 0.51 $} & \textbf{86.6 $\pm 0.44 $} \\
             \specialrule{1.5pt}{1pt}{1pt}
             & &  \multicolumn{3}{c}{\textbf{CRC-TP $\rightarrow$ LC25000}}\\Method  & Type & 1-shot & 5-shot & 10-shot\\
             \toprule
             ProtoNet \cite{prototypical_nets} & \multirow{5}{*}{Inductive} &  58.6 $\pm 0.49$ & 68.5 $\pm 0.36$ & 70.8 $\pm 0.32$\\
             MetaOpt \cite{lee2019meta} & & 49.2 $\pm 0.54$ & 72.9 $\pm 0.38$ & 78.5 $\pm 0.3$\\
             SimpleShot \cite{simpleshot} & & 62.1 $\pm 0.50 $ & 73.8 $\pm 0.36 $ & 76.6 $\pm 0.32$\\
             Distill \cite{tian2020rethinking} & & \textbf{63.2 $\pm 0.5 $} & 75.9 $\pm 0.37$ & 79.5 $\pm 0.33$\\
             Finetune \cite{closer_look} & & 62.1 $\pm 0.51$ & \textbf{79.2 $\pm 0.33$} & \textbf{83.7 $\pm 0.27$}\\
             \midrule
             MAML \cite{maml} & \multirow{3}{*}{Transductive} & 55.4 $\pm 0.51$ & 69.6 $\pm 0.41$ & 70.6 $\pm 0.36$\\
             LaplacianShot \cite{Laplacian} & & 62.4 $\pm 0.51$& 73.7 $\pm 0.36$ & 76.4 $\pm 0.32$\\
             TIM \cite{TIM}  &  &  \textbf{67.5 $\pm 0.62 $} & \textbf{80.9 $\pm 0.34 $} & \textbf{85.2 $\pm 0.28 $}\\
             \specialrule{1.5pt}{1pt}{1pt}
             & &  \multicolumn{3}{c}{\textbf{CRC-TP $\rightarrow$ BreakHis}}\\Method  & Type & 1-shot & 5-shot & 10-shot\\     
              \toprule
              ProtoNet \cite{prototypical_nets} & \multirow{5}{*}{Inductive} &  32.0 $\pm 0.48$ & 40.8 $\pm 0.49$ & 44.0 $\pm 0.47$\\
              MetaOpt \cite{lee2019meta} & & 28.3 $\pm 0.45$ & 42.5 $\pm 0.49$ & 49.1 $\pm 0.48$\\
              SimpleShot \cite{simpleshot} & & 36.2 $\pm 0.51 $ & 47.5 $\pm 0.51 $ & 51.8 $\pm 0.5$\\
              Distill \cite{tian2020rethinking}  & & 36.2 $\pm 0.5 $ & 50.2 $\pm 0.5 $ & 55.7 $\pm 0.48$\\
              Finetune \cite{closer_look} & & \textbf{36.4 $\pm 0.51$} & \textbf{56.9 $\pm 0.52$} & \textbf{66.7 $\pm 0.48$}\\
              \midrule
              MAML \cite{maml} & \multirow{3}{*}{Transductive} & 34.9 $\pm 0.52$ & 47.8 $\pm 0.52$ & 54.0 $\pm 0.51$\\
              LaplacianShot \cite{Laplacian} & & 36.5 $\pm 0.51$ & 47.9 $\pm 0.52$ & 52.3 $\pm 0.5$\\
              TIM \cite{TIM} &  &  \textbf{37.1 $\pm 0.59 $} & \textbf{57.9 $\pm 0.57 $} & \textbf{68.1 $\pm 0.51 $} \\
              \specialrule{1.5pt}{1pt}{1pt}
        \end{tabular}
        \label{tab:main_results}
    \end{table}
}

\newcommand*\ScalesTable{
    \begin{table}[h]
        \centering
        \small
        \caption{Test accuracy for different scales (magnifications).}
        \begin{tabular}{lcccccc}
            \toprule
            & & & & \multicolumn{3}{c}{CRC-TP(\textbf{20X}) $\rightarrow$ BreakHis}  \\
             Method  & Backbone & Image Size & Scale & 1-shot & 5-shot & 10-shot  \\
             \toprule
             \multirow{4}{*}{TIM\cite{TIM}} & \multirow{4}{*}{ResNet18} & \multirow{4}{*}{$84\times84$} & \textbf{40X} & 31.2 $\pm 0.46$ & 47.3 $\pm 0.48 $& 56.3 $\pm 0.46 $\\
             & & & \textbf{100X} & 29.0 $\pm 0.44$ & 44.1 $\pm 0.48$ & 52.8 $\pm 0.46$ \\
             & & & \textbf{200X} & 30.1 $\pm 0.46$ & 45.7 $\pm 0.47$ & 54.3 $\pm 0.46$ \\
             & & & \textbf{400X} & 29.8 $\pm 0.49$ & 43.6 $\pm 0.46$ & 51.4 $\pm 0.46$ \\
             \bottomrule
        \end{tabular}
        \label{tab:scales}
    \end{table}
}

\newcommand*\MainTableNCT{
    \begin{table}
        \centering
        \small
        \caption{Test accuracy in three different scenarios with considering NCT as the base dataset.}The backbone network in all the experiments is Resnet-18.
        \begin{tabular}{lcccc}
            \specialrule{1.5pt}{1pt}{1pt}
            & & \multicolumn{3}{c}{\textbf{NCT $\rightarrow$ CRC-TP}} \\
             Method & Type & 1-shot & 5-shot & 10-shot\\
             \toprule
             ProtoNet \cite{prototypical_nets} & \multirow{5}{*}{Inductive} &  36.2 $\pm 0.5$ & 50.4 $\pm 0.45$ & 54.6 $\pm 0.43$ \\
             MetaOpt \cite{lee2019meta} &  & 30.0 $\pm 0.45 $ & 42.7 $\pm 0.45 $ & 49.1 $\pm 0.44 $\\
             SimpleShot \cite{simpleshot} &  & 39.7 $\pm 0.54 $ & 55.6 $\pm 0.48 $ & 59.5 $\pm 0.44 $\\
             Distill \cite{tian2020rethinking} &  & \textbf{41.5 $\pm 0.53$} & 55.4 $\pm 0.47 $ & 60.0 $\pm 0.43 $\\
             Finetune \cite{closer_look} & & 39.7 $\pm 0.55$ & \textbf{56.3 $\pm 0.48$} & \textbf{62.2 $\pm 0.45$}  \\
             \midrule
             MAML \cite{maml} & \multirow{3}{*}{Transductive} & 36.1  $\pm 0.59 $ & 54.3 $\pm 0.5 $ & 60.4 $\pm 0.51 $\\
             LaplacianShot \cite{Laplacian} &  & 39.5 $\pm 0.54 $ & 55.4 $\pm 0.48 $ & 59.5 $\pm 0.44$\\
             TIM \cite{TIM} &  & \textbf{42.8 $\pm 0.64 $} & \textbf{58.1 $\pm 0.51 $} & \textbf{63.4 $\pm 0.46 $} \\
             \specialrule{1.5pt}{1pt}{1pt}
             & & \multicolumn{3}{c}{\textbf{NCT $\rightarrow$ LC25000}}\\Method  & Type & 1-shot & 5-shot & 10-shot\\
             \toprule
             ProtoNet \cite{prototypical_nets} & \multirow{5}{*}{Inductive} &  54.0 $\pm 0.57$ & 65.2 $\pm 0.41$ & 67.0 $\pm 0.37$\\
             MetaOpt \cite{lee2019meta} & & 53.4 $\pm 0.54 $ & 65.0 $\pm 0.38 $ & 70.2 $\pm 0.35 $\\
             SimpleShot \cite{simpleshot} & & \textbf{61.8 $\pm 0.64 $} & 74.4 $\pm 0.36 $ & 76.3 $\pm 0.32$\\
             Distill \cite{tian2020rethinking} & & 59.3 $\pm 0.56 $ & 71.4 $\pm 0.38$ & 75.1 $\pm 0.31$\\
             Finetune \cite{closer_look} & & 59.4 $\pm 0.61$ & \textbf{75.4 $\pm 0.34 $} & \textbf{79.1 $\pm 0.28 $}\\
             \midrule
             MAML \cite{maml} & \multirow{3}{*}{Transductive} &  55.3 $\pm 0.57$ & 74.7 $\pm 0.36$ & 81.7 $\pm 0.3$\\
             LaplacianShot \cite{Laplacian} & & 61.8 $\pm 0.63$& 74.3 $\pm 0.36$ & 76.4 $\pm 0.32$\\
             TIM \cite{TIM}  &  & \textbf{67.8 $\pm 0.74 $} & \textbf{80.8 $\pm 0.33 $} & \textbf{83.6 $\pm 0.28 $}\\
             \specialrule{1.5pt}{1pt}{1pt}
             & & \multicolumn{3}{c}{\textbf{NCT $\rightarrow$ BreakHis}}\\Method  & Type & 1-shot & 5-shot & 10-shot\\     
              \toprule
              ProtoNet \cite{prototypical_nets} & \multirow{5}{*}{Inductive} & 37.4 $\pm 0.54$ & 50.0 $\pm 0.55$ & 52.6 $\pm 0.53$\\
              MetaOpt \cite{lee2019meta} & & 29.7 $\pm 0.49 $ & 41.3 $\pm 0.5 $ & 48.9 $\pm 0.48 $\\
              SimpleShot \cite{simpleshot} & & 37.7 $\pm 0.53 $ & 50.6 $\pm 0.53 $ & 54.7 $\pm 0.51$\\
              Distill \cite{tian2020rethinking}  & & \textbf{38.2 $\pm 0.52 $} & 50.4 $\pm 0.51 $ & 55.9 $\pm 0.49$\\
              Finetune \cite{closer_look} & & 35.9 $\pm 0.49 $ & \textbf{55.1 $\pm 0.51 $} & \textbf{65.4 $\pm 0.48 $}\\
              \midrule
              MAML \cite{maml} & \multirow{3}{*}{Transductive} & 33.5 $\pm 0.47$ & 49.2 $\pm 0.52$ & 57.0 $\pm 0.51$\\
              LaplacianShot \cite{Laplacian} & & 37.7 $\pm 0.54$ & 50.6 $\pm 0.53$ & 54.7 $\pm 0.51$\\
              TIM \cite{TIM} &  & \textbf{38.4 $\pm 0.63 $} & \textbf{57.8 $\pm 0.55 $} & \textbf{66.8 $\pm 0.51 $} \\
              \specialrule{1.5pt}{1pt}{1pt}
        \end{tabular}
        \label{tab:main_results_nct}
    \end{table}
}

\newcommand*\BackboneTableNew{
	\begin{table} [h]
		\centering
		\caption{Comparison of different backbones. Note: DeiT-Tiny model \cite{touvron2020deit} has 2x fewer parameters compared to Resnet-18(5.5M vs 11M). Models with $^\ast$ are pre-trained on ImageNet dataset.}
		\resizebox{0.8\textwidth}{!}{
    		\begin{tabular}{lcccccc} 
    		    \toprule
    			Target Dataset & Method & Backbone & Image Size & 1-shot & 5-shot & 10-shot \\
    			\midrule
    			\multirow{7}{*}{NCT} & \multirow{4}{*}{TIM\cite{TIM}} & WRN-28-10 & $84\times84$ & 71.2 $\pm 0.86$ & 83.7 $\pm 0.49$ & 87.0 $\pm 0.43$ \\
    			& & DeiT-Tiny$^R$ & $224\times224$ & 48.9 $\pm 0.77$ & 63.3 $\pm 0.59$ & 64.8 $\pm 0.56$\\
    			& & DeiT-Tiny & $224\times224$ & 65.3 $\pm 0.86$ & 79.4 $\pm 0.54$ & 83.5 $\pm 0.45$\\
    			& & DeiT-Tiny$^\ast$ & $224\times224$ & \textbf{75.9 $\pm 0.85$} & \textbf{86.4 $\pm 0.47$} & \textbf{89.1 $\pm 0.39$}\\
    			\cmidrule(lr){3-7}
    			& \multirow{3}{*}{Finetune\cite{closer_look}} & WRN-28-10 & $84\times84$ & 64.9 $\pm 0.68$ & 81.2 $\pm 0.47$ & 85.0 $\pm 0.42$ \\
    			& & DeiT-Tiny & $224\times224$ & 58.9 $\pm 0.69$ & 76.8 $\pm 0.51$ & 81.2 $\pm 0.45$\\
    			& & DeiT-Tiny$^\ast$ & $224\times224$ & 66.5 $\pm 0.65$ & 83.6 $\pm 0.44$ & 87.3 $\pm 0.38$\\ \midrule
    			\multirow{7}{*}{LC25000} & \multirow{4}{*}{TIM\cite{TIM}} & WRN-28-10 & $84\times84$ & 67.1 $\pm 0.58$ & 81.8 $\pm 0.36$ & 87.0 $\pm 0.27$\\
    			& & DeiT-Tiny$^R$ & $224\times224$ & 58.7 $\pm 0.61$ & 69.4 $\pm 0.41$ & 73.4 $\pm 0.35$\\
    			& & DeiT-Tiny & $224\times224$ & 66.7 $\pm 0.67$ & 81.9 $\pm 0.36$ & 86.3 $\pm 0.28$\\
    			& & DeiT-Tiny$^\ast$ & $224\times224$ & \textbf{75.7 $\pm 0.53$} & \textbf{87.9 $\pm 0.31$} & \textbf{91.4 $\pm 0.24$}\\
    			\cmidrule(lr){3-7}
    			& \multirow{3}{*}{Finetune\cite{closer_look}} & WRN-28-10 & $84\times84$ & 61.4 $\pm 0.52$ & 79.5 $\pm 0.35$ & 84.7 $\pm 0.28$\\
    			& & DeiT-Tiny & $224\times224$ & 61.9 $\pm 0.55$ & 80.2 $\pm 0.36$ & 84.9 $\pm 0.29$\\
    			& & DeiT-Tiny$^\ast$ & $224\times224$ & 69.8 $\pm 0.45$ & 85.0 $\pm 0.29$ & 89.7 $\pm 0.24$\\ \midrule
    			\multirow{7}{*}{BreakHis} & \multirow{4}{*}{TIM\cite{TIM}} & WRN-28-10 & $84\times84$ & 39.1 $\pm 0.6 $ & 60.0 $\pm 0.57$ & 70.9 $\pm 0.52$\\
    			& & DeiT-Tiny$^R$ & $224\times224$ & 32.4 $\pm 0.51$ & 41.2 $\pm 0.48$ & 45.0 $\pm 0.48$\\
    			& & DeiT-Tiny & $224\times224$ & 36.2 $\pm 0.59$ & 55.3 $\pm 0.55$ & 65.2 $\pm 0.51$\\
    			& & DeiT-Tiny$^\ast$ & $224\times224$ & \textbf{39.9} $\pm 0.6 $ & \textbf{60.4} $\pm 0.58$ & \textbf{71.1} $\pm 0.54$\\
    			\cmidrule(lr){3-7}
    			& \multirow{3}{*}{Finetune\cite{closer_look}} & WRN-28-10 & $84\times84$ & 38.7 $\pm 0.54 $ & 59.9 $\pm 0.52$ & 69.9 $\pm 0.49$\\
    			& & DeiT-Tiny & $224\times224$ & 36.3 $\pm 0.51$ & 56.1 $\pm 0.5$ & 65.6 $\pm 0.49$\\
    			& & DeiT-Tiny$^\ast$ & $224\times224$ & 38.8 $\pm 0.5 $ & 59.8 $\pm 0.52$ & 70.0 $\pm 0.49$\\
    			\bottomrule
    		\end{tabular}
    	}
		\label{tab:backbones}
	\end{table}
}

\newcommand*\HistologyDatasets{
    \begin{table}[]
        \centering
        \caption{Histology Datasets.}
        \resizebox{\textwidth}{!}{%
        {\renewcommand{\arraystretch}{1.0}
        \setlength\tabcolsep{2pt}
        \begin{tabular}{@{}l|c|c|c|c|c|l|l@{}}
            \toprule
            \textbf{Dataset} & \multicolumn{1}{l|}{\textbf{\#Train}} & \multicolumn{1}{l|}{\textbf{\#Val}} & \multicolumn{1}{l|}{\textbf{\#Test}} & \multicolumn{1}{l|}{\textbf{\#Class}} & \multicolumn{1}{l|}{\textbf{Size}} & \textbf{Types} & \textbf{Other Information} \\ 
            \midrule
            \textbf{\begin{tabular}[c]{@{}l@{}}\href{https://doi.org/10.1016/j.media.2019.05.010}{BACH}  \cite{aresta2019bach}\\ (Microscopy)\end{tabular}} & 400 & - & 100 & 4 & 2048 x 1536 & \begin{tabular}[c]{@{}l@{}}Normal\\ Benign\\ In Situ Carcinoma\\ Invasive Carcinoma\end{tabular} & \begin{tabular}[c]{@{}l@{}}Breast cancer\\ Different patients in the test test \\ Available upon request (\href{https://iciar2018-challenge.grand-challenge.org/Download/}{ICIAR 2018}) \\ Format: .tiff\\ 95\% Classification Accuracy (100 test images) (\href{https://www.sciencedirect.com/science/article/pii/S0933365719306621}{paper})\end{tabular} \\
            \midrule
            \textbf{\href{https://doi.org/10.5281/zenodo.1214456}{NCT-CRC-HE-100K} \cite{kather100_data}} & 100,000 & 7,180 & - & 9 & 224 x 224 & \begin{tabular}[c]{@{}l@{}}Adipose\\ Background \\ Debris \\ Lymphocytes \\ Mucus \\ Smooth Muscle \\ Normal Colon Mucosa \\ Cancer-associated Stroma\\ Colorectal Adenocarcinoma Epithelium\end{tabular} & \begin{tabular}[c]{@{}l@{}}Colorectal cancer\\ Non-overlapping image patches\\Ready to use (\href{https://doi.org/10.5281/zenodo.1214456}{zenodo.org}) \\ Format: .tiff\\ 98.7\% Classification Accuracy in an internal testing set (70\% 15\% 15\% splits from 100K)\\ 94.3\% Classification Accuracy in an external testing set (7180 test images) (\href{https://journals.plos.org/plosmedicine/article?id=10.1371/journal.pmed.1002730}{paper})\end{tabular} \\ 
            \midrule
            \textbf{\href{https://www.kaggle.com/kmader/colorectal-histology-mnist}{Colorectal Histology}  \cite{kather2016multi}} & \begin{tabular}[c]{@{}c@{}}5,000  \\ (+ 10 large)\end{tabular} & - & - & 8 & \begin{tabular}[c]{@{}c@{}}150 x 150\\ (5000 x 5000)\end{tabular} & \begin{tabular}[c]{@{}l@{}}Tumour Epithelium\\ Simple Stroma\\ Complex Stroma\\ Immune Cells\\ Debris\\ Normal Mucosal Glands\\ Adipose Tissue\\ Background\end{tabular} & \begin{tabular}[c]{@{}l@{}}Colorectal cancer\\ 8 different types of tissue (tumor/non tumor)\\ No train/val/test splits\\ Ready to use (\href{https://www.tensorflow.org/datasets/catalog/colorectal_histology}{Tensorflow-datasets})\\ Format: .tiff\\95.4\% Classification Accuracy (90\% train, 10\% test (10-fold cross-validation)) (\href{https://ieeexplore.ieee.org/abstract/document/9108668}{paper})\end{tabular} \\ 
    
            \midrule
            \textbf{\href{https://web.inf.ufpr.br/vri/databases/breast-cancer-histopathological-database-BreakHis/}{BreakHis} \cite{spanhol2015dataset}} & 7,909 & - & - & \begin{tabular}[c]{@{}c@{}}8 \\ (4 Malignant\\ + 4 Benign)\end{tabular} & 700 x 460 & \begin{tabular}[c]{@{}l@{}}Adenosis (A)\\ Fibroadenoma (F)\\ Phyllodes Tumor (PT)\\ Tubular Adenona (TA)\\ Carcinoma (DC)\\ Lobular Carcinoma (LC)\\ Mucinous Carcinoma (MC)\\ Papillary Carcinoma (PC)\end{tabular} & \begin{tabular}[c]{@{}l@{}}Breast cancer\\ No train/val/test splits\\ Different magnifying factors (40X, 100X, 200X, 400X)\\ Available upon request (\href{https://web.inf.ufpr.br/vri/databases/breast-cancer-histopathological-database-BreakHis/}{BreakHis}) \\ Format: .png\\ 98.8\% Classification Accuracy (80\% train 20\% test) (\href{https://www.sciencedirect.com/science/article/abs/pii/S0378437119319995}{BreastNet Paper})\end{tabular} \\ 
            \midrule
            \textbf{\href{https://jgamper.github.io/PanNukeDataset/}{PanNuke} \cite{gamper2019pannuke}} & 2656 & 2523 & 2722 & 19 & 256 x 256 & \begin{tabular}[c]{@{}l@{}}Adrenal\_gland, Bile-duct, Bladder, Breast, \\ Cervix, Colon, Esophagus, HeadNeck, \\ Kidney, Liver, Lung, Ovarian, Pancreatic, \\ Prostate, Skin, Stomach, Testis, Thyroid, \\ Uterus\end{tabular} & \begin{tabular}[c]{@{}l@{}}Includes nuclei segmentation labels across 19 different tissue types\\ Three folds used as train/val/test interchangeably\\ Ready to use (\href{https://jgamper.github.io/PanNukeDataset/}{PanNuke}) \\ Format: .npy\end{tabular} \\
            \midrule
            \begin{tabular}[c]{@{}l@{}} \href{https://www.sciencedirect.com/science/article/pii/S2352914820300757} {\textbf{Databiox paper}} \cite{bolhasani2020histopathological} \\ \href{https://databiox.com/}{\textbf{Databiox data}}\end{tabular} & 922 & - & - & 3 & \begin{tabular}{c} 2100 x 1574 (24-bit) \\ 1276 x 956 (24-bit) \end{tabular}   & \begin{tabular}[c]{@{}l@{}} Grade I \\ Grade II \\ Grade III \end{tabular} & \begin{tabular}[c]{@{}l@{}}Breast cancer\\ Available upon request (\href{https://databiox.com/}{DataBiox}) \\ Images in .jpeg \\ 4 levels of magnif. x 4, x10, x20, x40  \end{tabular}   \\  
            \midrule
            \begin{tabular}[c]{@{}l@{}} \href{https://arxiv.org/pdf/1912.12142.pdf} {\textbf{LC25000 paper}} \cite{borkowski2019lung}\\ \href{https://github.com/tampapath/lung_colon_image_set}{\textbf{LC25000 data}}\end{tabular} & 25'000 & - & - & 5 &  768 x 768  & \begin{tabular}[c]{@{}l@{}} Benign lung \\ Lung adenocarcinoma \\ Squamous cell carcinomas \\ Benign colon \\ Colon adenocarcinoma \end{tabular} & \begin{tabular}[c]{@{}l@{}}Lung and colon cancer \\ 1250 original images $\rightarrow$ 25'000 after augm. \\98.56\% Classification Accuracy (75\% train 25\% test) (\href{https://ieeexplore.ieee.org/document/9298990}{paper}) \\ format: jpg \end{tabular}   \\ 
            
            \midrule 
            \begin{tabular}[c]{@{}l@{}} \href{https://www.researchgate.net/publication/344188411_Multiplex_Cellular_Communities_in_Multi-Gigapixel_Colorectal_Cancer_Histology_Images_for_Tissue_Phenotyping} {\textbf{CRC-TP paper}} \cite{javed2020multiplex} \\ \href{https://warwick.ac.uk/fac/cross_fac/tia/data/crc-tp}{\textbf{CRC-TP data}}\end{tabular} & 280'000 & - & - & 7 &  150 x 150  & \begin{tabular}[c]{@{}l@{}} Tumor \\ Stroma \\ Complex Stroma \\ Muscle \\ Debris \\ Inflammatory \\ Benign \end{tabular} & \begin{tabular}[c]{@{}l@{}}Colorectal Cancer \\ Ready to use(\href{https://warwick.ac.uk/fac/sci/dcs/research/tia/data/crc-tp}{warwick})  \\ 280k patches in total, two differnet splits: $\rightarrow$ \\ Train: 70\%  - Test: 30\%  \\ Train: 14 patients - Test: 6 patients\\91\% F1 Score (patch-level separation between training and testing splits)\\84\% F1 Score (patient-level separation between training and testing splits) (\href{https://www.sciencedirect.com/science/article/pii/S136184152030061X}{paper})  \end{tabular}   \\ 

            \midrule
            \textbf{\begin{tabular}[c]{@{}l@{}}\href{https://doi.org/10.5281/zenodo.3373439}{No Name} \\ {[}from TCGA{]}\end{tabular}} & 1,608,060 & - & - & 32 & 256 x 256 &  & \begin{tabular}[c]{@{}l@{}}Different types of cancer\\ \textcolor{red}{Restricted Access (Paper is under revision.)}\\ No train/val/test splits\\ Format: .jpg\end{tabular} \\       
            \midrule
           
            \href{https://figshare.com/articles/dataset/BreCaHAD_A_Dataset_for_Breast_Cancer_Histopathological_Annotation_and_Diagnosis/7379186}{\textbf{BReCaHad}} \cite{aksac2019brecahad} & 162 & - & - & 6 &  1360 x 1024 & \begin{tabular}[c]{@{}l@{}}Mitosis \\ apoptosis \\ tumor nuclei \\ non tumor nuclei \\ tubule \\ non tubule \end{tabular} & \begin{tabular}[c]{@{}l@{}}Breast cancer\\x,y coordinates indicate the centroid of the annotated object\\ Ready to use (\href{https://figshare.com/articles/dataset/BreCaHAD_A_Dataset_for_Breast_Cancer_Histopathological_Annotation_and_Diagnosis/7379186}{BReCaHad}) \\ Images in .png \\ Annotations in .json  \end{tabular}   \\ 
            \midrule
            \textbf{\begin{tabular}[c]{@{}l@{}}\href{https://panda.grand-challenge.org/data/}{PANDA}\end{tabular}} & 11000 & - & 1000 & 6 & WSI & \begin{tabular}[c]{@{}l@{}}background (non tissue) or unknown(0)\\ stroma(connective tissue, non-epithelium tissue)(1)\\ healthy (benign) epithelium(2)\\ cancerous epithelium (Gleason 3)\\ cancerous epithelium (Gleason 4)\\cancerous epithelium (Gleason 5)\end{tabular} & \begin{tabular}[c]{@{}l@{}}prostate cancer\\ \textbf{MICCAI 2020 challenge}\\ Two data providors:\\ 1. Radboud: Prostate glands are individually labelled(6 classes) \\ 2. Karolinska: Regions are labelled(3 classes)\\ Sample Code (\href{https://www.kaggle.com/drhabib/ens-xie-2fold-drhb-igor-ru-se50-ru-efnet-5}{2nd Place}) \\ 94\% Accuracy(private test set)\\ Format: .tiff\end{tabular} \\
            
            \bottomrule
        \end{tabular}%
        \label{tab:histologydatasets}
        }
        
        }
    \end{table}
}

\newcommand*\ChallengesTable{
    \begin{table}[]
        \centering
        \caption{Challenges on histology datasets.}
        \resizebox{\textwidth}{!}{%
        {\renewcommand{\arraystretch}{1.0}
        \setlength\tabcolsep{2pt}
        \begin{tabular}{@{}l|l|l|l@{}}
            \toprule
            \textbf{Challenge} & \textbf{Goal} & \textbf{Top performance} & \textbf{Other information} \\ \midrule
            \textbf{\begin{tabular}[c]{@{}l@{}}BACH\\ (\href{https://iciar2018-challenge.grand-challenge.org}{website})\\ (\href{https://doi.org/10.1016/j.media.2019.05.010}{paper})\end{tabular}} & \begin{tabular}[c]{@{}l@{}}Typical supervised setting\\ Part A: classification of breast microscopy images\\ Part B: pixel-wise labelling of breast whole-slide images\end{tabular} & \begin{tabular}[c]{@{}l@{}}        91.3 Classification Acc., model: Inception V3 (\href{https://www.sciencedirect.com/science/article/pii/S1046202319300349?casa_token=1ZMgbHmMVVMAAAAA:YrZdz5np3DS8ItMa2467SlpmvhUPiPsDjTKxkEtO-HS3toMnYyCOpmcfjeQuVdcbHABj72jvOClN}{paper})\\ 96.1 (transfer learning from other microscopy images)(\href{https://www.mdpi.com/2079-9292/9/3/445}{paper})(Custom model)\end{tabular} & \begin{tabular}[c]{@{}l@{}}2018\\ 4 Classes\\400 train imgs\\100 test imgs\end{tabular} \\
            \midrule
            \textbf{\begin{tabular}[c]{@{}l@{}}MoNuSAC 2020
            \\ (\href{https://monusac-2020.grand-challenge.org/}{\textbf{website}}) \\(\href{https://www.researchgate.net/publication/339227864_Multi-organ_Nuclei_Segmentation_and_Classification_Challenge_2020} {\textbf{paper}})  \\ \end{tabular}} & \begin{tabular}[c]{@{}l@{}}Multi-organ Nuclei Segmentation and Classification\\\\Unequal number of instances in the training dataset which\\encourages participants to develop algorithms for learning\\ from imbalanced classes in a few shot learning paradigm.\\\\ No domain shift  \end{tabular} & \begin{tabular}[c]{@{}l@{}} 61.2 \% PQ 
            (=weighted average of the class-specific panoptic
            quality) (\href{https://www.sciencedirect.com/science/article/pii/S136184152030061X}{paper})\\\end{tabular} & \begin{tabular}[c]{@{}l@{}}\\2020\\\\ 31,411 hand-annotated nuclei instances including:\\
            14,539 epithelial cells,\\ 15,654 lymphocytes,\\ 587 macrophages,\\
            and 631 neutrophils\\\\The testing set will be created from the\\ patients not included in
            the training set\\\\The testing data will be prepared using the similar\\ protocol
            as adopted for creating the challenge training data\end{tabular} \\           
            \midrule
            
            \textbf{\begin{tabular}[c]{@{}l@{}}PatchCamelyon\\ (\href{https://patchcamelyon.grand-challenge.org/}{website})\\ (\href{https://arxiv.org/abs/1806.03962}{paper})\end{tabular}} & \begin{tabular}[c]{@{}l@{}}Typical supervised setting\\Classification of Lymph node metastases\end{tabular} & 100\% AUC & \begin{tabular}[c]{@{}l@{}}2018\\ Binary classification\\ 327,680 images\\ extracted from Camelyon16  \\75/12.5/12.5 train/val/test \\ 96x96\end{tabular} \\
            
            \midrule
            \textbf{\begin{tabular}[c]{@{}l@{}}NCT-CRC-HE-100K\\ (\href{https://doi.org/10.5281/zenodo.1214456}{website})\\\end{tabular}} & \begin{tabular}[c]{@{}l@{}}Typical supervised setting\\Classification of colorectal cancer\end{tabular} & \begin{tabular}[c]{@{}l@{}}        98.7\% Classification Acc. in an internal testing set (70\% 15\% 15\% splits from 100K)\\ 94.3\% Classification Acc. in an external testing set (7180 test images) (\href{https://journals.plos.org/plosmedicine/article?id=10.1371/journal.pmed.1002730}{paper})\\ model: alexnet, googlenet, resnet50, squeezenet, and vgg19\end{tabular} & \begin{tabular}[c]{@{}l@{}}2018\\ 9 Classes\\100K train\\7180 val\end{tabular} \\    
            
            \midrule
            \textbf{\begin{tabular}[c]{@{}l@{}}Colorectal Histology\\ (\href{https://www.kaggle.com/kmader/colorectal-histology-mnist}{website})\\(\href{https://www.nature.com/articles/srep27988}{paper})\end{tabular}} & \begin{tabular}[c]{@{}l@{}}Typical supervised setting\\Classification of colorectal cancer\end{tabular} & \begin{tabular}[c]{@{}l@{}} 95.4\% Classification Accuracy (90\% train, 10\% test (10-fold cross-validation)) (\href{https://ieeexplore.ieee.org/abstract/document/9108668}{paper}) \\ model: SqueezeNet, MobileNet, ResNet, DenseNet\end{tabular} & \begin{tabular}[c]{@{}l@{}}2016\\ 8 Classes\\5K train\end{tabular} \\          
      
            \midrule
            \textbf{\begin{tabular}[c]{@{}l@{}}BreakHis\\ (\href{https://web.inf.ufpr.br/vri/databases/breast-cancer-histopathological-database-BreakHis/}{website})\\(\href{https://ieeexplore.ieee.org/document/7312934}{paper})\end{tabular}} & \begin{tabular}[c]{@{}l@{}}Typical supervised setting\\Classification of breast cancer\end{tabular} & \begin{tabular}[c]{@{}l@{}} 98.8\% Classification Accuracy (80\% train 20\% test) (\href{https://www.sciencedirect.com/science/article/abs/pii/S0378437119319995}{ paper})(Custom CNN model)\end{tabular} & \begin{tabular}[c]{@{}l@{}}2016\\ 8 Classes\\7,909 train\end{tabular} \\            
      
            \midrule
            \textbf{\begin{tabular}[c]{@{}l@{}}LC25000\\(\href{https://github.com/tampapath/lung_colon_image_set}{\textbf{website}})\\(\href{https://arxiv.org/pdf/1912.12142.pdf} {\textbf{paper}}) \\ \end{tabular}} & \begin{tabular}[c]{@{}l@{}}Typical supervised setting\\Classification of Lung and colon cancer\end{tabular} & \begin{tabular}[c]{@{}l@{}} 98.56\% Classification Accuracy (75\% train 25\% test) (\href{https://ieeexplore.ieee.org/document/9298990}{paper})\\model: Desnset-121, ResNet-50\end{tabular} & \begin{tabular}[c]{@{}l@{}}2019\\ 5 Classes\\25,000 train\end{tabular} \\ 
            \midrule
            \textbf{\begin{tabular}[c]{@{}l@{}}CRC-TP\\ (\href{https://warwick.ac.uk/fac/sci/dcs/research/tia/data/crc-tp}{\textbf{website}}) \\(\href{https://www.researchgate.net/publication/344188411_Multiplex_Cellular_Communities_in_Multi-Gigapixel_Colorectal_Cancer_Histology_Images_for_Tissue_Phenotyping} {\textbf{paper}})  \\ \end{tabular}} & \begin{tabular}[c]{@{}l@{}}Typical supervised setting\\Classification of Colorectal Cancer \end{tabular} & \begin{tabular}[c]{@{}l@{}} 91\% F1 Score (patch-level separation between training and testing splits)\\84\% F1 Score (patient-level separation between training and testing splits) (\href{https://www.sciencedirect.com/science/article/pii/S136184152030061X}{paper})\\Train: 70\%  - Test: 30\%\\model: custom model\end{tabular} & \begin{tabular}[c]{@{}l@{}}2020\\ 7 Classes\\280,000 train\end{tabular} \\

            \bottomrule
            \end{tabular}%
            \label{tab:challenges}
        }
        }
    \end{table}
}

\newcommand*\DatasetsFig{
\begin{figure}
    \begin{subfigure}{3.3cm}
    \centering\includegraphics[width=3.3cm, frame]{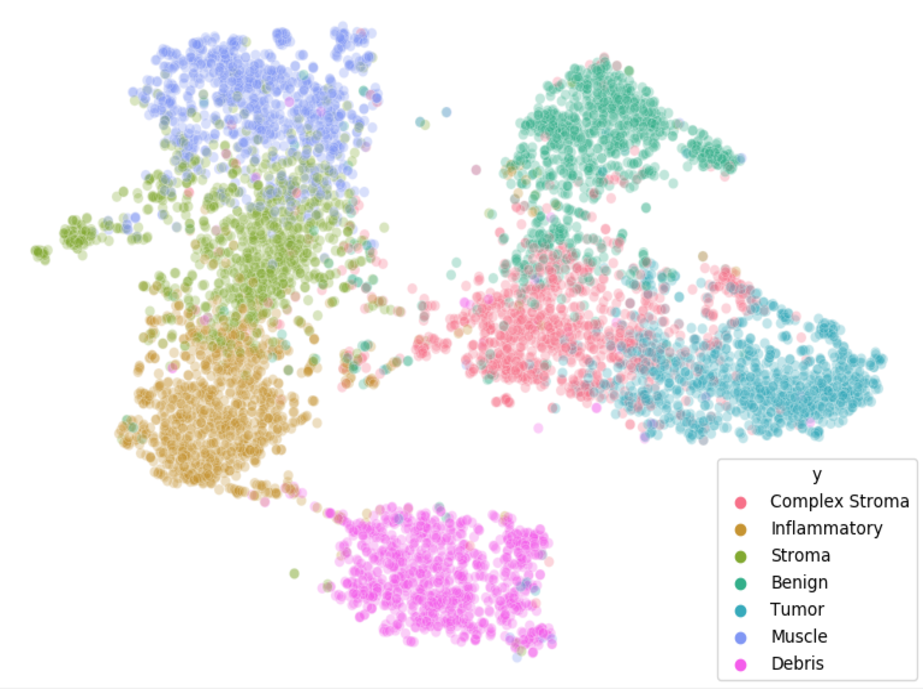}
    \caption{CRC-TP}
    \end{subfigure}
    \begin{subfigure}{3.3cm}
    \centering\includegraphics[width=3.3cm, frame]{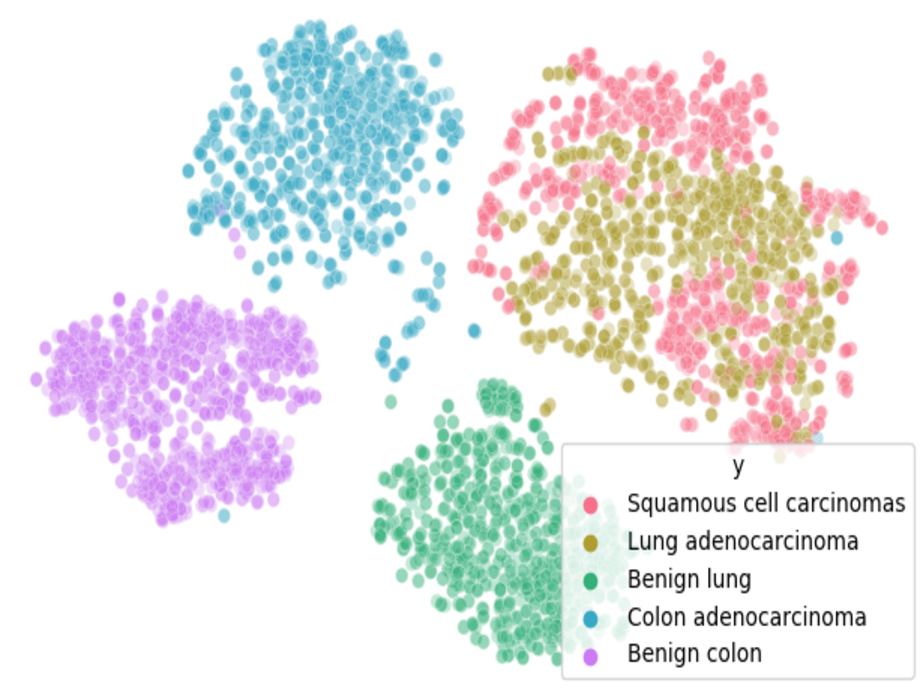}
    \caption{LC25000}
    \end{subfigure}
    \begin{subfigure}{3.3cm}
    \centering\includegraphics[width=3.3cm, frame]{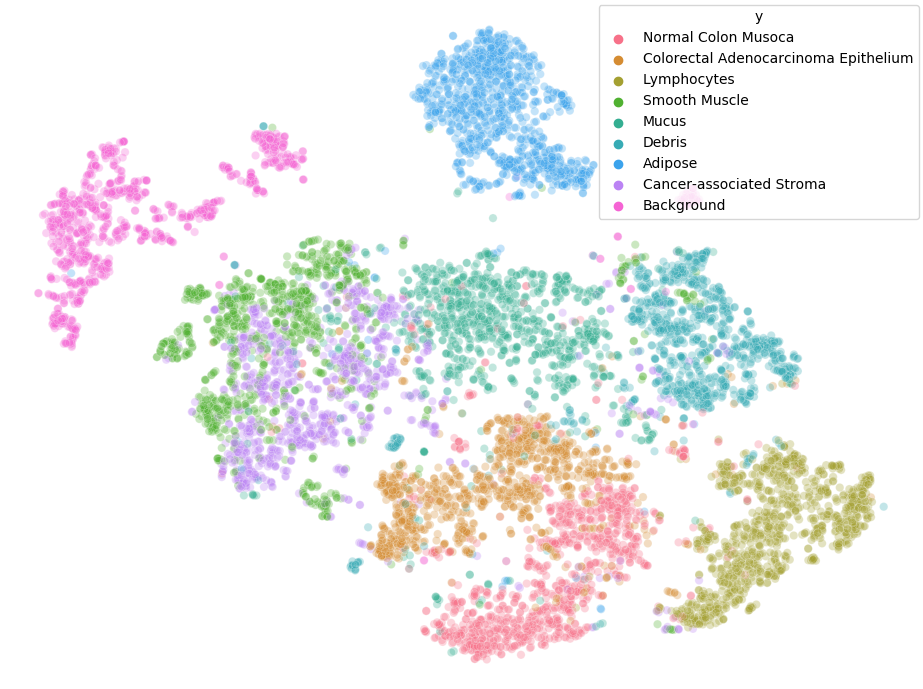}
    \caption{NCT}
    \end{subfigure}
    \begin{subfigure}{3.3cm}
    \centering\includegraphics[width=3.3cm, frame]{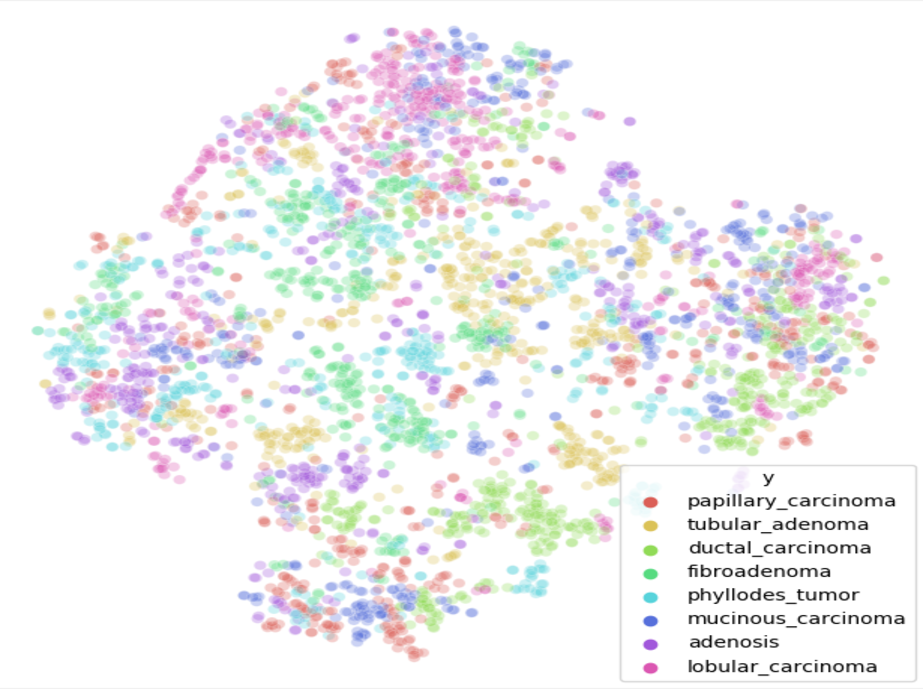}
    \caption{BreakHis}
    \end{subfigure}
    \caption{t-SNE maps of CRC-TP, LC25000, NCT, and BreakHis datasets. These figures depict similarity and dissimilarity between classes of each dataset. Figure (d) shows the difficulty of classification task on BreakHis dataset since classes are not well clustered.}
    \label{fig:t-sne}
\end{figure}
}
\newcommand*\DomainShiftFig{
\begin{wrapfigure}{r}{0.6\textwidth}
    \begin{subfigure}{2.5cm}
    \centering\includegraphics[width=2.5cm]{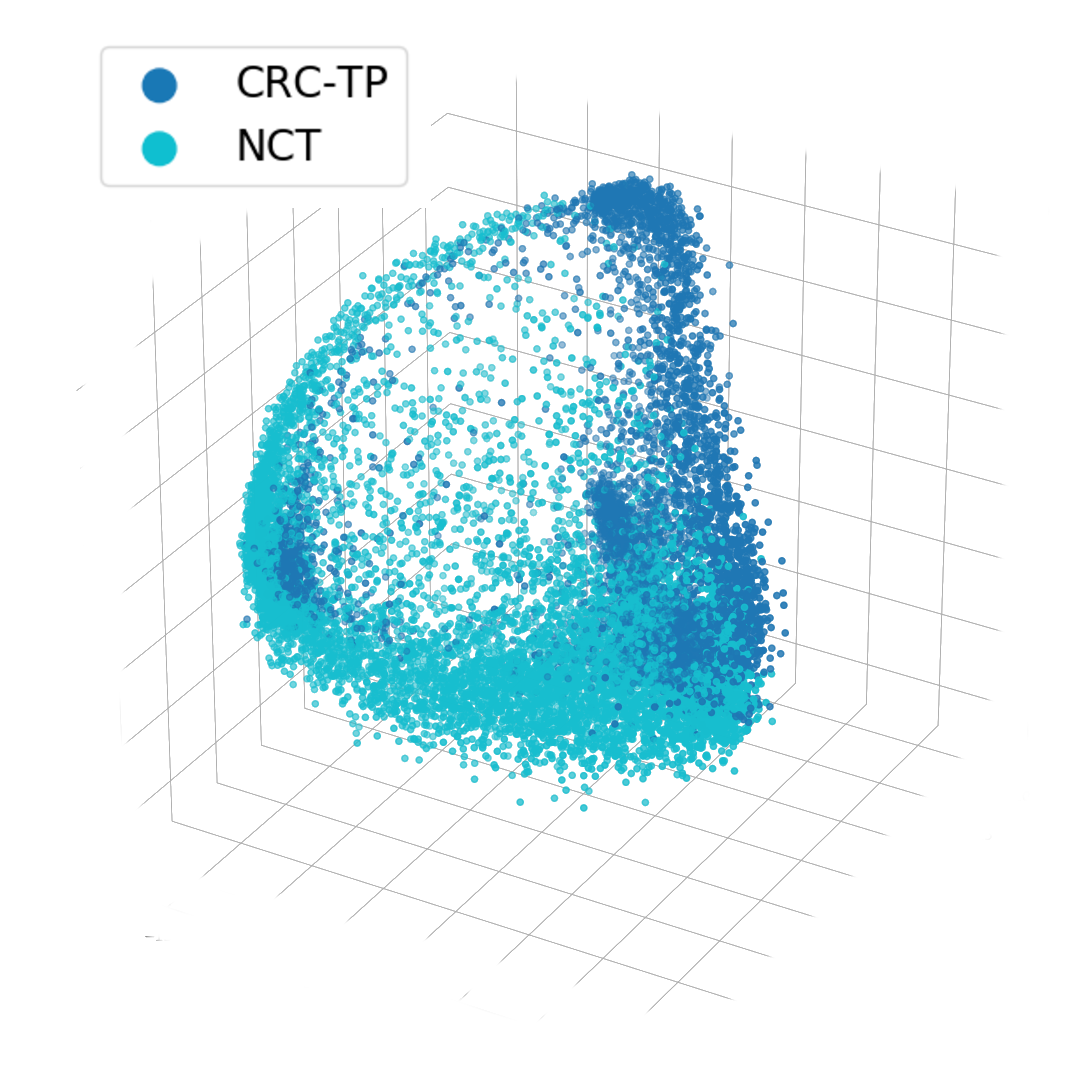}
    \caption{\tiny{CRC-TP/NCT}}
    \end{subfigure}
    \begin{subfigure}{2.5cm}
    \centering\includegraphics[width=2.5cm]{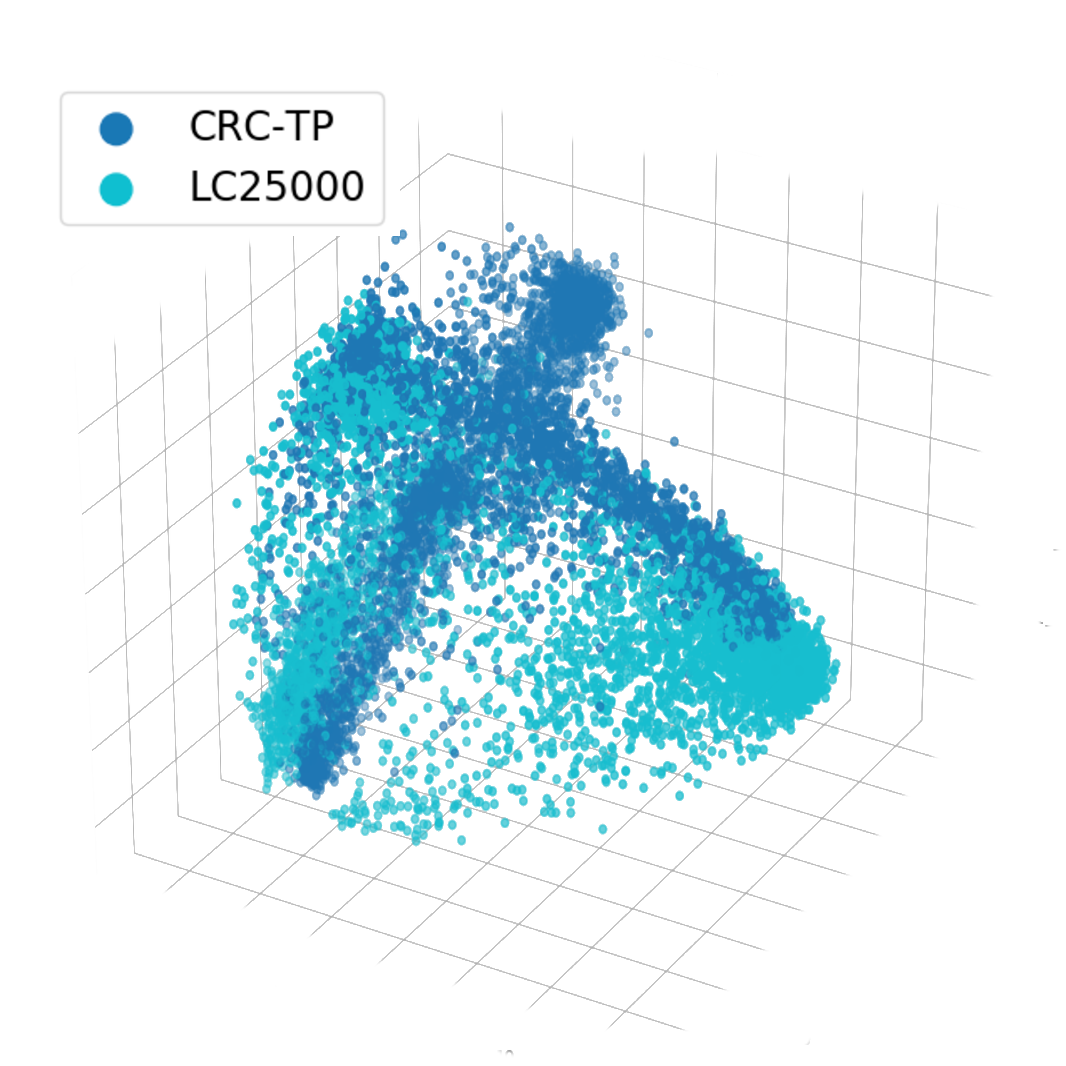}
    \caption{\tiny{CRC-TP/LC25000}}
    \end{subfigure}
    \begin{subfigure}{2.5cm}
    \centering\includegraphics[width=2.5cm]{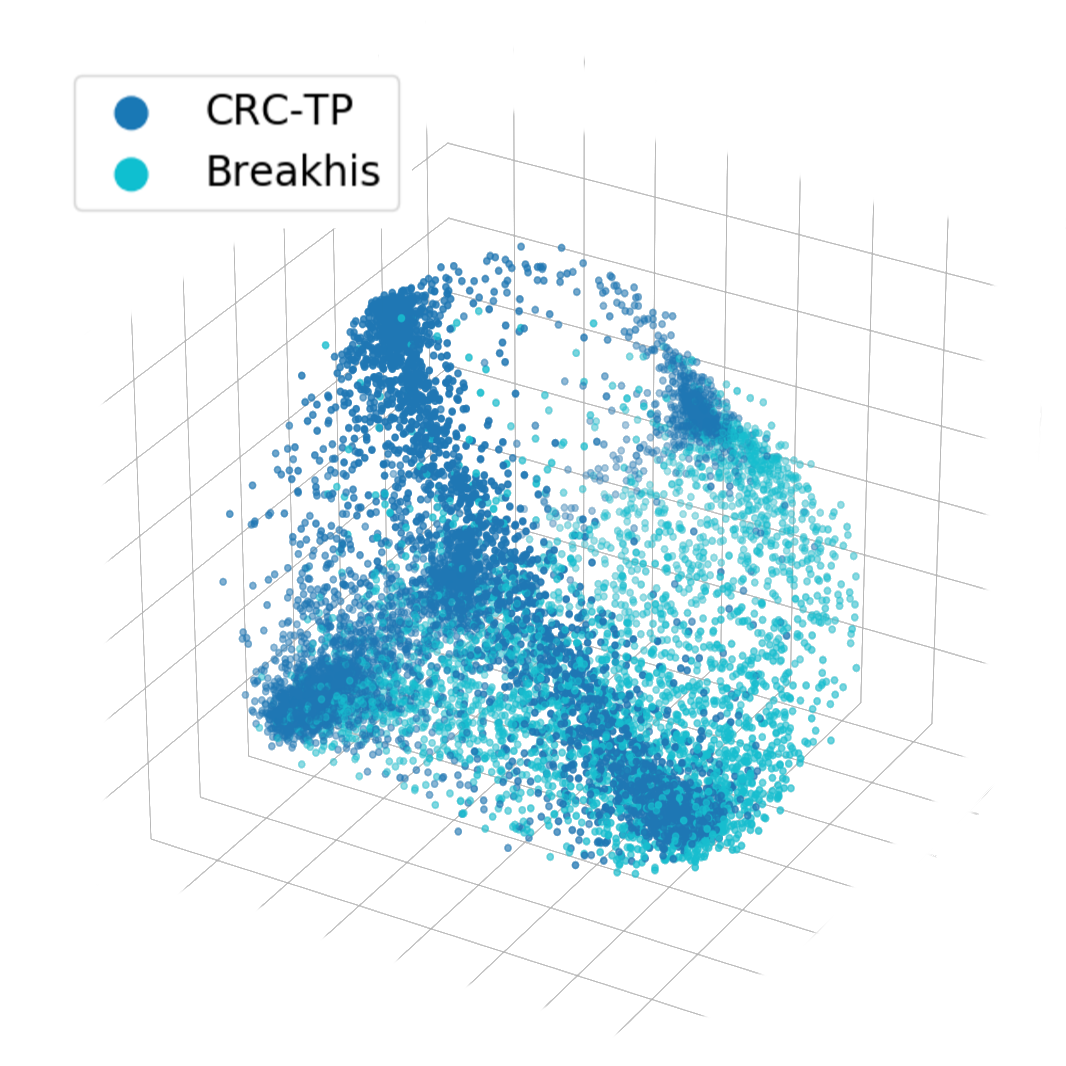}
    \caption{\tiny{CRC-TP/Breakhis}}
    \end{subfigure}
    \caption{PCA Visualization of domain shift between datasets. In all figures, dark blue points represent samples of CRC-TP dataset, and light blue color in the (a), (b), and (c) represent NCT, LC25000, and BreakHis, respectively. }
    \label{fig:pca}
\end{wrapfigure}
}

\newcommand*\TaskFig{
\begin{figure}[t!]
    \centering
    \includegraphics[width=\textwidth]{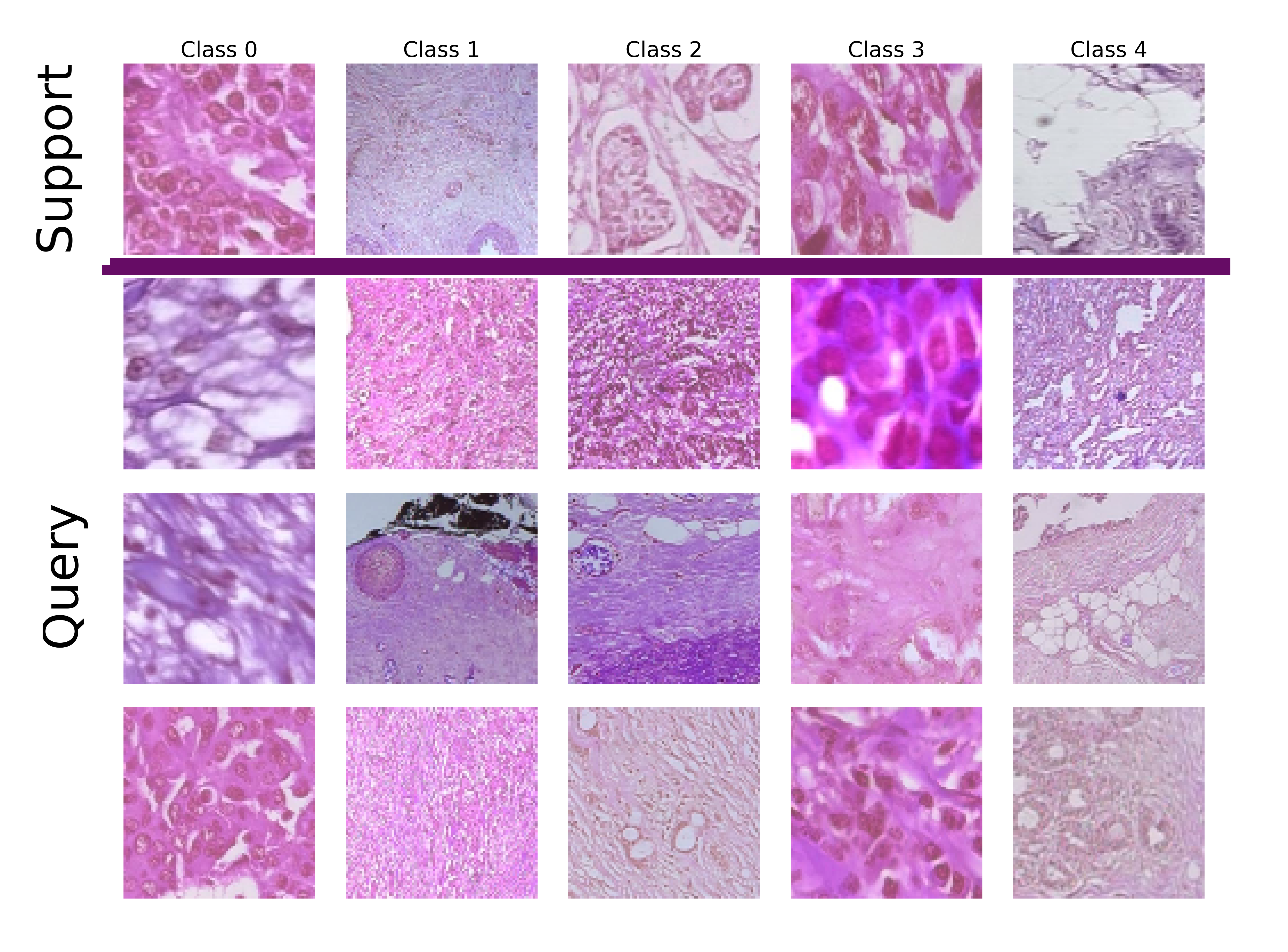}
    \caption{Example of a 5-way 1-shot 15-query,  task. Images of the first row represent support samples from 5 different classes of BreakHis. The 15 images in the last three rows represent query images sampled from BreakHis dataset.}
    \label{fig:taskexample}
\end{figure}
}

\newcommand*\CRCFig{
\begin{figure}[t!]
    \centering
    \includegraphics[width=\textwidth]{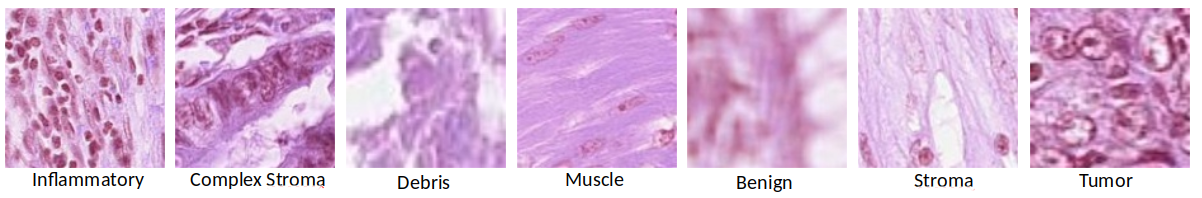}
    \caption{Examples from different classes of CRC-TP dataset}
    \label{fig:crctpexample}
\end{figure}
}

\newcommand*\NCTFig{
\begin{figure}[t!]
    \centering
    \includegraphics[width=\textwidth]{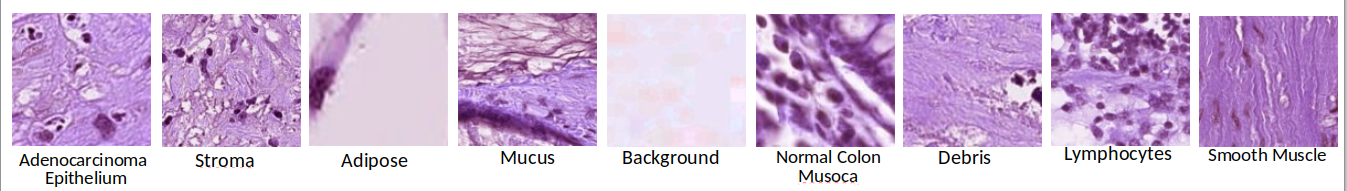}
    \caption{Examples from different classes of NCT dataset}
    \label{fig:nctexample}
\end{figure}
}

\newcommand*\LCFig{
\begin{figure}[t!]
    \centering
    \includegraphics[width=\textwidth]{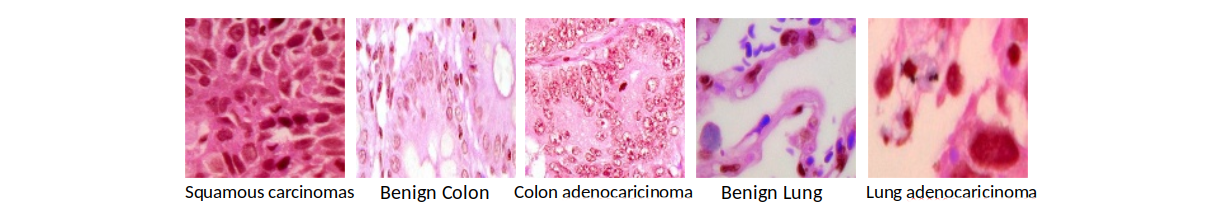}
    \caption{Examples from different classes of LC25000 dataset}
    \label{fig:lcexample}
\end{figure}
}

\newcommand*\BHFig{
\begin{figure}[t!]
    \centering
    \includegraphics[width=\textwidth]{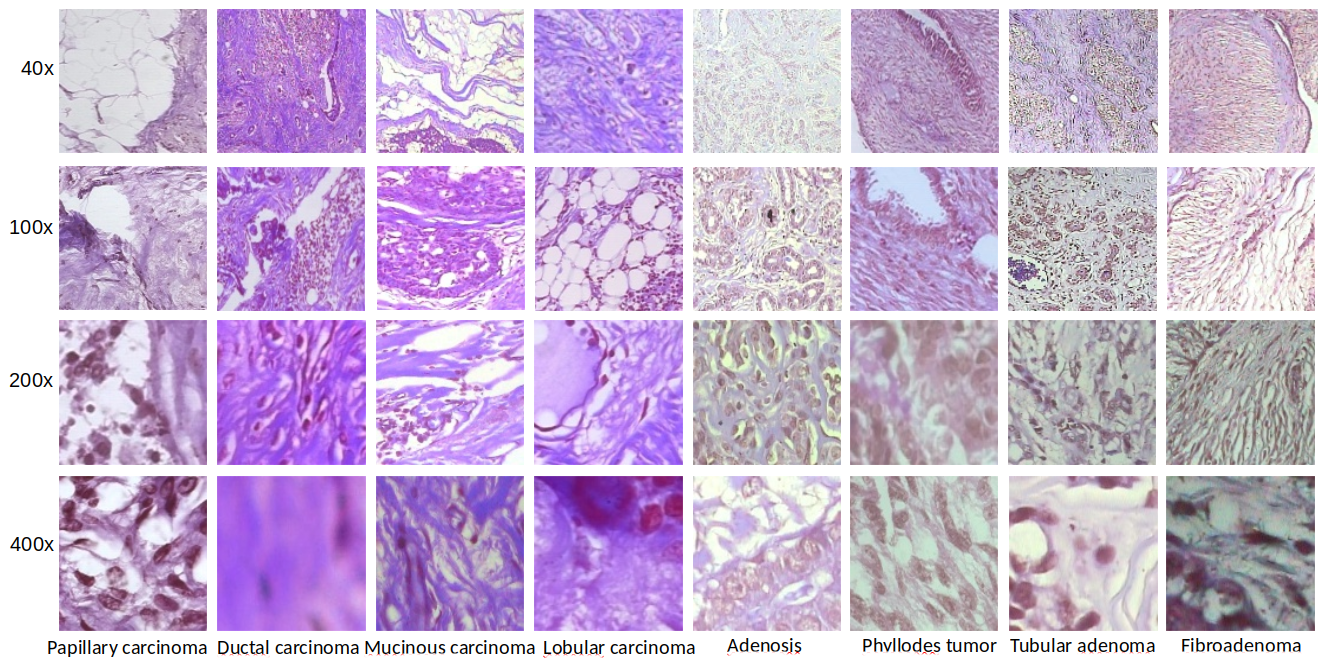}
    \caption{Examples from different classes and magnifications of BreakHis dataset}
    \label{fig:breakhisexample}
\end{figure}
}

\newcommand*\FHISTFig{
\begin{figure}[t!]
    \centering
    \includegraphics[width=\textwidth]{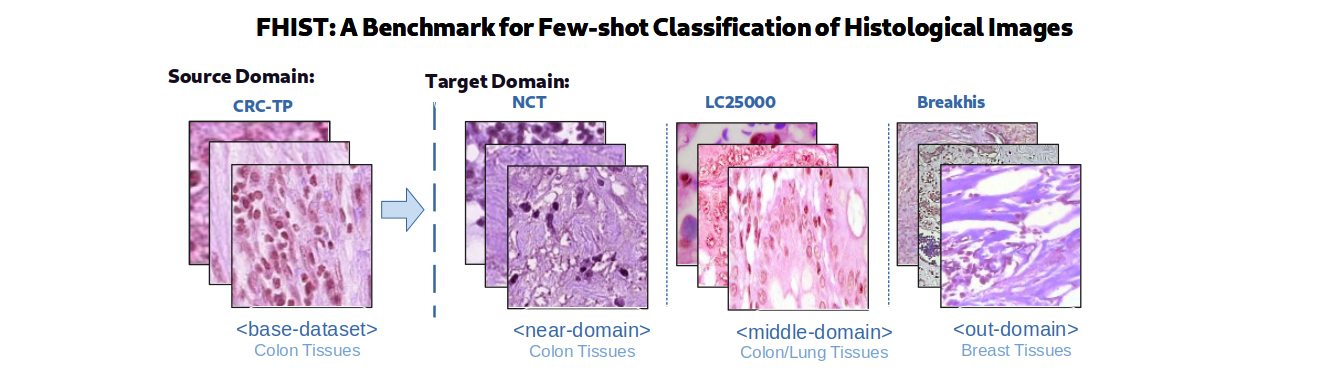}
    \caption{Our proposed few-shot histology benchmark, FHIST. In our main experiments, we consider CRC-TP(colorectal tissues) as the base training source dataset. We define three different adaptation scenarios based on the domain shift between the source and the target histology datasets: near-domain(NCT: colorectal tissues), middle-domain(LC25000: colorectal and lung tissues), and out-domain(BreakHis: breast tissues).}
    \label{fig:fhist}
\end{figure}
}

\usepackage[nonatbib]{neurips_data_2021}

\usepackage[utf8]{inputenc} 
\usepackage[T1]{fontenc}    
\usepackage{hyperref}       
\usepackage{url}            
\usepackage{booktabs}       
\usepackage{amsfonts}       
\usepackage{nicefrac}       
\usepackage{microtype}      
\usepackage{xcolor}         
\usepackage{ bbold }
\usepackage{amsmath}
\usepackage{amssymb}
\usepackage{wrapfig}

\usepackage{subcaption}
\usepackage{graphicx}
\usepackage[export]{adjustbox}
\usepackage[english]{babel}
\usepackage{multirow}
\usepackage{ dsfont }

\newcommand{\neardomain}{near-domain}
\newcommand{\middledomain}{middle-domain}
\newcommand{\outdomain}{out-domain}

\hypersetup{
    colorlinks=true,
    linkcolor=blue,
    filecolor=magenta,      
    urlcolor=cyan,
}

\title{FHIST: A Benchmark for Few-shot Classification of Histological Images}

\author{%
  Fereshteh Shakeri\thanks{Equal contributions. Corresponding authors: \{fereshteh.shakeri.1, malik.boudiaf.1\}@etsmtl.net} \\
  ÉTS Montréal, CRCHUM\\
  \And
  Malik Boudiaf$^*$ \\
  ÉTS Montréal \\
  \And
  Sina Mohammadi \\
  University of Twente\\
  \AND
  Ivaxi Sheth\\
  ÉTS Montréal, Mila \\
  \And
  Mohammad Havaei \\
  Imagia Inc, Montréal \\
  \And
  Ismail Ben Ayed \\
  ÉTS Montréal, CRCHUM \\
  \And
  Samira Ebrahimi Kahou \\
  ÉTS Montréal, Mila, CIFAR \\
}

\begin{document}

\maketitle

\begin{abstract}

Few-shot learning has recently attracted wide interest in image classification, but almost all the current public benchmarks are focused on 
natural images. The few-shot paradigm is highly relevant in medical-imaging applications due to the scarcity of labeled data, as annotations are expensive
 and require specialized expertise. However, in medical imaging, few-shot learning research is sparse, limited to private data sets and is at its early 
 stage. In particular, the few-shot setting is of high interest in histology (study of diseased tissues with microscopic images) due to the diversity and fine granularity of cancer related 
 tissue classification tasks, and the variety of data-preparation techniques, which results in covariate shifts in the inputs and disparities 
 in the labels.
This paper introduces a highly diversified public benchmark, gathered from various public datasets, for few-shot histology data classification. 
We build few-shot tasks and base-training data with various tissue types, different levels of domain shifts stemming from various cancer sites, and 
different class-granularity levels, thereby reflecting realistic scenarios. 
We evaluate the performances of state-of-the-art few-shot learning methods on our benchmark, and observe that simple fine-tuning and regularization methods 
achieve significantly better results than the popular meta-learning and episodic-training paradigm.
Furthermore, we introduce three scenarios based on the domain shifts between the source and target histology data: (i) \neardomain, (ii) \middledomain~and (iii) \outdomain. 
Our experiments display the potential of few-shot learning in histology classification, with state-of-art few shot learning methods  approaching the supervised-learning baselines in the near-domain setting. In our out-domain setting, for 5 way 5 shot, the best performing method reaches 60\% accuracy. 
We believe that our work could help in building realistic evaluations and fair comparisons of few-shot learning methods and will further encourage research in the few-shot paradigm.
 Our code and few-shot histology tasks are publicly available~\footnote{\href{https://github.com/mboudiaf/Few-shot-histology}{https://github.com/mboudiaf/Few-shot-histology}}.

\end{abstract}

\section{Introduction}

Cancer remains one of the leading causes of death in North America despite scientific breakthroughs in the last decade~\footnote{\href{https://www.cdc.gov/}{https://www.cdc.gov/}}. Cancers are classified either by tissue types (named as histological type) or by the body location (primary site) where the cancer first appeared \cite{cancer_classification}. There are hundreds of cancer types from a histological perspective \cite{cancer_classification}, making it a difficult and diversified classification problem. Histology slides taken from Whole Slide Image (WSI) \cite{pantanowitz2010digital} scanners
are pivotal in cancer diagnosis and staging \cite{madabhushi2009digital}, as they provide a comprehensive view of the tissues affected by the disease, \cite{hipp2011pathology} which can further determine treatment options. Cancers are predominantly diagnosed by pathologists, who analyze WSIs to determine various tissue types. Analyzing these WSIs manually is a massive workload for pathologists, which increases the turnaround reporting time substantially. Furthermore,
in realistic clinical settings, cancer-related tissue classification tasks are highly diverse, i.e., correspond to different cancer sites. Even within the same cancer site, the tasks 
might come at different class-granularity levels; see Table \ref{tab:datasets}. Also, evaluations by histologists might be subject to high inter-observer variability. Hence, automating tissue-type classification in histology images is of high value clinically, and has triggered a large body of works within the image processing and machine learning communities, e.g.,  \cite{komura2018machine, petushi2006large, qureshi2008adaptive, tabesh2007multifeature, bilgin2007cell}, just to list a few.  

Deep neural networks \cite{goodfellow2016deep} are currently dominating both the machine learning and computer vision literature, including a breadth of applications in
medical image analysis 
\cite{litjens2017survey, razzak2018deep, dimitriou2019deep, srinidhi2020deep}. However, to achieve good performances, deep learning models require large annotated datasets.
The few-shot learning paradigm attempts to tackle this challenge, and has recently triggered substantial research efforts, with a large body of new methods, mostly evaluated on 
natural image benchmarks. Given a large number of training examples for some “base” classes, few-shot learning algorithms aim to recognize “novel” classes, unseen during base 
training, with a limited number of labeled examples per novel class. Such algorithms are often evaluated on large sets of few-shot tasks (or episodes), each containing a 
labeled set (the support set) and a set of unlabeled points for testing (the query set). 

Few-shot learning is highly relevant in medical imaging due to the scarcity of labeled data, as annotations are expensive and require specialized expertise. 
However, in this application domain, the few-shot literature remains sparse, confined to private data sets and is at an early stage compared to the very large body of few-shot 
works focused and evaluated on standard natural-image benchmarks. Specifically, in histology imaging, the few-shot paradigm is of high clinical interest due to 
the high diversity and fine granularity of cancer related tissue classification tasks encountered in realistic settings. Another challenge with histopathology data is the 
variety of data-preparation techniques, such as different coloring or annotation, which results in covariate shifts in the input space and disparities in the label space. 
Therefore, training deep learning models on large data sets may not be always possible, especially for machine-learning experts who lack domain knowledge. 

Despite these challenges, the vast majority of a large body of machine-learning research
carried out on histology data is focused on the fully supervised setting, i.e., assuming: (i) a large amount of labeled data is available at training and (ii) the testing and training data correspond to the same classes and are drawn from the same distributions. The literature on few-shot histology classification is very limited and confined to in-house data sets, binary classification and specific cancer sites, which do not reflect the high diversity and fine granularity of realistic tissue classification tasks. Also, it is not clear
to what extent the state-of-the-art few-shot methods designed for natural images are applicable to histological data classification. The lack of a 
standard public benchmark for few-shot histology makes it difficult to evaluate these methods quantitatively.

We introduce a highly diversified public benchmark, gathered from various public datasets, for few-shot histology data classification. We build few-shot tasks and base-training data with various cancer related tissue types, different domain-shift levels resulting from various cancer sites and different class-granularity levels of the classification, which encourages research in this important application area and enables fair comparisons of few-shot learning methods in the context of histology. We report comprehensive evaluations of the performances of recent state-of-the-art inductive and transductive few-shot learning methods, which were initially designed for natural images, on our benchmark. 
We observe that the very popular and sophisticated meta-learning methods are not well suited for this new benchmark.
Simple fine-tuning and regularization techniques at inference, following a standard training on base data, can achieve significantly better results than popular and intensively researched meta-learning and episodic-training methods.
We further examine three scenarios based on the domain shifts between the source and target histology data. Our experiments point to the potential and challenges of few-shot learning in histology classification. On the one hand, in the near-domain setting, where classes of two datasets are semantically related but have different levels of granularity, few-shot performances approached those of supervised-learning baselines. On the other hand, in the challenging out-domain setting, with different class types belonging to different cancers, the best-performing method achieves 60\% accuracy on average using five labeled examples. 

\section{Related Work}
\subsection{Few-shot learning}
In what follows, we discuss few-shot learning approaches divided into three main categories: metric-based meta-learning, optimization-based meta-learning, and generative and augmentation based approaches. Then, we discuss a few recent works focused on applying few-shot methods to medical-imaging problems.

\textbf{Metric-based meta-learning. }This category of methods tackles few-shot classification by “learning to compare”. The core idea is to learn feature representations that preserve the class neighborhood structure. In other words, these methods aim to enforce the features from the same class to be closer to each other than the features from different classes. Specifically, Koch et al. \cite{koch2015siamese} train Siamese neural networks~\cite{chopra2005learning} to pull apart images belonging to different categories, while reducing discrepancy between images of the same category, via a metric learning loss. Vinyals et al. \cite{vinyals2016matching} introduced Matching Networks to classify a test image by adopting a weighted nearest neighbor classifier, implemented with an attention mechanism to encode a full context embedding of the support samples. Snell et al. \cite{snell2017prototypical} proposed Prototypical Networks, which represents each category in the embedding space by the mean of the support samples (i.e., the class prototype). In this method, the classification is performed using the distances between the query samples and the prototypes. To boost the performances, Oreshkin et al \cite{oreshkin2018tadam} introduced some modifications to Prototypical Networks, including task-conditioning, metric scaling and auxiliary task co-training. Instead of employing distances between embeddings, Sung et al. \cite{sung2018learning} introduced a relation network for learnable metric comparison.  

\textbf{Optimization-based meta-learning.} This class of algorithms addresses few-shot classification by “learning to learn”, and attempt to optimize the model parameters, so as to tackle new tasks effectively. A number of works focused on learning good model initialization so that, with a small number of gradient updates, classification for novel classes could be learnt with a few labeled examples \cite{finn2017model,finn2018probabilistic,rusu2018meta}. Another line of works aims to learn an optimizer. For example, Ravi \& Larochelle \cite{ravi2016optimization} replaced the stochastic gradient decent optimizer with an LSTM-based meta-learner. Munkhdalai \& Yu \cite{munkhdalai2017meta} used a weight-update mechanism with an external memory. These approaches could achieve fast adaptation with a few training examples per novel classes. However, the recent studies in \cite{closer_look,guo2020broader} showed that the performances of these approaches may degrade substantially when there are strong domain shifts between the base and novel classes. In another work, Lee et al \cite{lee2019meta} proposed MetaOptNet, which uses linear classifiers (e.g., SVM or linear regression) instead of 
nearest-neighbor classifiers. They showed that the former achieves better performances than the latter since it can learn better 
class boundaries.

\textbf{Generative and augmentation based approaches.} This class of methods tackles few-shot classification by producing more samples from a few training examples, using either generative models to synthesize new data points, data-augmentation strategies \cite{lim2019fast}, or additional examples from external data. For instance, in \cite{hariharan2017low}, a generator learned from base-class data is used to transfer variance to the novel classes. The generators in \cite{antoniou2017data} adopted generative adversarial networks (GAN)~\cite{goodfellow2014generative} to transfer the style. In \cite{wang2018low}, to improve the classification accuracy, the generator is directly integrated into a meta-learning algorithm rather than explicitly specifying what to transfer. Reed et al. \cite{reed2017few} performed few-shot class density estimation using autoregressive models, in combination with neural attention, in which examples are synthesized by a sequential process. Some works focused on the role of label and attribute semantics as additional information to facilitate few-shot learning \cite{chen2019multi,schwartz2019baby,yu2017semantic}.

All the above discussed few-shot methods focused on the standard setting where the base classes and
novel classes are from the same domain, and are evaluated on natural images using the current standard benchmarks 
for few-shot classification: Omniglot \cite{lake2011one}, CUB \cite{wah2011caltech}, miniImageNet \cite{vinyals2016matching}, tieredImageNet \cite{ren2018meta} and CIFAR-FS \cite{bertinetto2018meta}. In contrast to these standard natural-image benchmarks, our benchmark focuses on histology images for classifying various tissue types and includes datasets from different domains (i.e., different cancer sites), along with an investigation as to how few-shot methods perform when the gap between histology-classification domains becomes large.

\textbf{Few-shot learning in medical image analysis}

A few recent works tackled few-shot learning in specific medical-imaging areas \cite{ mahajan2020meta, chen2021momentum, ouyang2020self, medela2019few, yu2021location,teng2021few}. In \cite{ mahajan2020meta}, for instance, two few-shot learning methods, namely Reptile \cite{nichol2018reptile} and Prototypical Networks \cite{snell2017prototypical}, were applied to skin-disease identification. In \cite{ chen2021momentum}, a two-stage framework was adopted for COVID-19 diagnosis from chest CT images. In the first stage, expressive feature representations are captured by an encoder trained on publicly available lung datasets using contrastive learning. In the second stage, this pre-trained encoder was adopted in a few-shot learning paradigm using the popular Prototypical Networks method \cite{snell2017prototypical}. In \cite{ouyang2020self}, a few-shot semantic segmentation framework was introduced to eliminate the need for labels during training. In \cite{yu2021location}, few-shot medical image segmentation was performed by adopting a prototype-based method, called location-sensitive local prototype network, which leverages spatial priors. In \cite{medela2019few}, a deep Siamese neural network was trained to transfer knowledge from a dataset containing colon tissue to another dataset containing colon, lung and breast tissue. Finally, \cite {teng2021few} proposed a few-shot learning algorithm based on Prototypical Networks, for classifiying lymphatic metastasis of lung carcinoma from Whole Slide Images (WSIs).

Despite the high clinical relevance of few-shot classification of histology images, the literature on the subject is, to our knowledge, very limited \cite{medela2019few,teng2021few} and confined to specific cancer sites, in-house data sets and binary classification, which do not reflect the high diversity and fine granularity of cancer related tissue classification tasks encountered in realistic settings. For instance, \cite{medela2019few} used a private data set and is limited to binary classification (cancerous/non-cancerous).  
To fill this gap, we introduce a highly diversified public benchmark, with various cancer sites, tissue types and granularity levels of the classification tasks.

\subsection{Domain adaptation}
Domain adaptation approaches have been intensively investigated in the recent years. The purpose is to transfer knowledge 
from a source domain to a target one, by effectively reducing the domain shifts \cite{choi2019pseudo,chen2019progressive,xie2018learning,sankaranarayanan2018generate,ghifary2016deep,ganin2015unsupervised}. Ghifary et al. \cite{ghifary2016deep} proposed to reconstruct data in the target domain to encode useful features from the domain. Inspired by GANs, several methods employed adversarial training to reduce the discrepancy between the feature distributions of the source and target domains, which resulted in learning domain-invariant features \cite{ganin2015unsupervised, chen2019progressive,  xie2018learning, choi2019pseudo}. In such methods, it is assumed that the classes of the training and testing sets are identical. Unlike these methods, in our study, some classes in the source and target domains are disjoint, which can be viewed as a cross-domain few-shot learning problem. Additionally, in our study, only a few examples are available in the target domain, a major difference with the assumptions of the existing domain-adaptation literature.

\subsection{Cross-domain Few-shot Learning}

Cross-domain Few-shot Learning addresses the problem where source and target domains have different labels, with a shift between the two domains. The recent study in Chen et al. \cite{ closer_look} examined cross-domain few-shot learning by investigating the performance of existing few-shot learning methods in this setting, using natural-image data sets of different domains and class granularity levels. 
The authors of \cite{phoo2021STARTUP} investigated a self-supervised strategy, which adapts feature representations trained on source tasks from ImageNet \cite{imagenet} to extremely different target domains, including medical chest X-rays radiology images \cite{wang2017chestx}. Their proposed method was very promising on non-medical data, but little improvement was 
observed in the case of chest X-rays. Tseng et al. \cite{ tseng2020cross} proposed to apply feature-wise transformation layers to different metric-based meta-learning approaches to improve their generalization capability. Koniusz et al. \cite{ koniusz2018museum} proposed a new dataset called Open Museum Identification Challenge (Open MIC), which limits the cross-domain setting to images of items in museum galleries. Guo et al. \cite{ guo2020broader} proposed a benchmark called Broader Study of Cross-Domain Few-Shot Learning (BSCD-FSL), which covers a large spectrum of image types including radiology images, dermatology images, and satellite images. 

\subsection{The main differences between the proposed benchmark and the existing few-shot learning benchmarks}

Despite these recent efforts devoted to cross-domain few-shot learning, to the best of our knowledge, there is no standard public benchmark focused on few-shot histology image classification, and which reflects the diversity and granularity of such a task in realistic clinical settings. To fill this gap, we propose FHIST benchmark to facilitate quantitative evaluation of  few-shot histology image classification methods.
The following  points clarify the differences between our new benchmark and the recent few-shot benchmarks in \cite{guo2020broader, closer_look}, in terms of the definitions of the domain-shift settings, interpretations of the results and evaluated few-shot techniques (transductive vs. inductive):

$~\bullet~$ We have considered three different scenarios based on the types of domain shifts between the source and the 
target histology data: near-domain, middle-domain, and out-domain. The near-domain setting is fundamentally different from 
those examined in \cite{closer_look, guo2020broader}, and corresponds to realistic histology settings, as it deals with 
semantically related classes but different levels of class granularity. This setting showed relatively high performances of state-of-the-art methods, approaching the supervised learning baselines. The results in this near-domain setting points to the relevance and potential impact of few-shot learning in histology. 

$~\bullet~$ In the out-domain setting, our findings are, to some extent, in line with \cite{guo2020broader}. However, \cite{guo2020broader} considers extreme domain shifts with the source and the target datasets coming from completely different domains (e.g. Natural images vs X-Ray images) which yielded poor performances of few-shot methods, approaching those of random classifiers (less than 30\% in 5-way tasks). For instance, in \cite{guo2020broader}, the best-performing method for Imagenet $\rightarrow$ ChestX 5-shot 5-way setting resulted in 27\% accuracy. This is different from our out-domain setting, which yielded significantly better performances. 
The best performing method in our CRC-TP $\rightarrow$ BreakHis 5-way 5-shot setting resulted in 60\% accuracy. We believe that, even in our out-domain setting, this 60\% performance is a good starting point to further encourage research in few-shot learning for histology image classification. 

$~\bullet~$ The transductive few-shot setting has recently attracted significant attention in few-shot learning, with a large body of recent works and performances that are significantly better than the inductive setting. The studies in \cite{closer_look, guo2020broader} focused on the inductive setting only. We have conducted experiments for both transductive and inductive settings.

\section{Proposed Benchmark}
\subsection{The data sources} \label{ch:data-sources}
In this section, we introduce FHIST(Figure~\ref{fig:fhist}), our proposed benchmark for few-shot classification of histological images. FHIST is composed of several histology datasets, namely  CRC-TP \cite{javed2020multiplex}, LC25000 \cite{borkowski2019lung}, BreakHis \cite{spanhol2015dataset}, and  NCT-CRC-HE-100K \cite{kather100_data}.  Throughout the rest of the paper, we refer to this dataset as ``NCT'' for ease of reference. \autoref{tab:datasets} describes details of different datasets used in this benchmark. In the last column of this table, we provide the results of the previous supervised methods trained on the whole training split of each dataset. These histology datasets consist of different tissue types and different organs. We consider each tissue type as a class label with one-hot encoding in the classification task. Figure~\ref{fig:t-sne} presents t-SNE maps~\cite{van2008visualizing} of each dataset. Different colors designate different tissue types. 
\FHISTFig
FHIST reflects a real-world few-shot learning problem as it encourages learning of models which can generalize across data from different domains using few labeled examples. Due to the difficulty of collecting large amount of labeled data, learning data-efficient models is very desirable for histological image classification.
This motivated us to introduce the FHIST benchmark to encourage research on learning transferable models using few labeled examples. 
Please refer to Appendix \ref{sec:datasets} for detailed description of each dateset included in our benchmark.

\DatasetsTable
\DatasetsFig

\subsection{The Suggested Train-Test Splits}

We evaluate few-shot learning under various degrees of domain shift. To that end, we consider
\DomainShiftFig
CRC-TP which is the largest dataset in the benchmark (containing only colorectal images) as our {\it base training} source dataset and construct three adaptation scenarios ($i$) 
\neardomain, in which NCT \cite{kather100_data} (another colon histology dataset) is taken as the target test dataset. While NCT and CRC-TP both contain images from the same organ (colon), the images are collected from different institutions but overlap on some of the tissue types ($ii$) \middledomain~in which LC25000 \cite{borkowski2019lung} (containing images from lung and colon) is taken as the target dataset and ($iii$) \outdomain~in which BreakHis \cite{spanhol2015dataset}--a dataset containing images from breast, an entirely different organ from that of the base training dataset--is taken as the target dataset.  Figure~\ref{fig:pca} shows the covariate shift between the source dataset (dark blue) and each of the target datasets (light blue).

\section{Methods}
\subsection{Few-shot classification problem formulation} \label{ch:few-shot}
Let $D_{\text{base}}$ be a large dataset 
of labeled examples often referred to as source dataset with label space $\mathcal{Y}_{\text{base}}$. We refer to classes from the source dataset as base classes.
The goal of few-shot classification is to produce a model that can leverage the knowledge learned from $D_{\text{base}}$ to learn from very few examples on new set of classes.
Training usually consists of two phases: pre-training phase on $D_{\text{base}}$ and  meta-test phase on an unseen target dataset $D_{\text{test}}$ with new set of classes $\mathcal{Y}_{\text{test}}$ such that $\mathcal{Y}_{\text{base}} \cap \mathcal{Y}_{\text{test}} = \emptyset$.  
For evaluation (meta-test), we use {\it episodic} sampling where a series of tasks are constructed from $D_{\text{test}}$.  For each task, we construct a  support set $\mathcal{S} := \{({\bf x}_i, y_i)\}_{i=1}^{|NK|}$ by randomly sampling $N$ classes from $\mathcal{Y}_{\text{test}}$. For each sampled class we randomly sample $K$ examples, resulting into an N-way K-shot classification task. Every task is accompanied by a query set for evaluation denoted by $\mathcal{Q}$, sampled from the same classes.  

\subsection{Inductive methods}
Inductive setting assumes complete separation between the labeled support set and the unlabeled query set. When adapting to a task, the model only learns from the support set and no information from the query set is provided to the model. Below, we detail four inductive baselines reproduced in our experiments.

\textbf{Finetune}
We follow the fine-tune procedure as Baseline++~\cite{closer_look}. In the pre-training phase, a feature extractor $f_{\phi}$ is trained on $D_{\text{base}}$ and $\mathcal{Y}_{\text{base}}$. In the fine-tune phase, $f_{\phi}$ is frozen and a new classifier is trained on the labels of the sampled task. Instead of a linear classifier, for every input example ${\bf x}_i$, the model computes a class similarity score between  $f_{\phi}({\bf x}_i)$ and weight vectors corresponding to each class. The likelihood of each class is achieved by applying softmax on the class similarity scores. 

\textbf{Distill}
We follow the procedure introduced in \cite{tian2020rethinking}. During the pre-training phase, a feature extractor is trained using all classes from $\mathcal{Y_{\text{base}}}$ as a single task. The loss is augmented with self-distillation from previous iterations of the same model. In the adaptation phase, the feature extractor is fixed and a linear classifier is trained on the extracted features for every task.

\textbf{Prototypical Networks}
Introduced by Snell et al.~\cite{snell2017prototypical}, Prototypical Networks is a meta-learning method that uses episodic sampling of tasks (see~\ref{ch:few-shot}) in the meta-training phase. At every task, the input examples are projected into a metric space through $f_\phi$. The goal of meta-learning is to optimize $\phi$ across tasks such that distance between examples in the embedding are semantically meaningful. Given the support set $\mathcal{S}$ and $f_\phi$, prototypical networks constructs a prototype for each class. Query examples from $\mathcal{Q}$ are classified based on the nearest prototype in the embedding space. 

\textbf{MetaOptNet}
MetaOptNet~\cite{lee2019meta} is a meta-learning method that uses episodic sampling in the pre-training (meta-training) phase.  Similar to ProtoNets, the feature extractor is frozen at meta-test phase. However, unlike ProtoNets, it uses differentiable quadratic programming solver such as SVM as the base learner.

\subsection{Transductive methods}
In the transductive setting, we assume the model has access to both the support set and the query set, although, only the support set contains labels. While adapting to a task, some information about the query set is integrated in the learning process; either the batch statistics~\cite{finn2017model}, or data level information in the form of semi-supervised learning~\cite{dhillon2019baseline, TIM}. 

\textbf{MAML}
MAML~\cite{finn2017model} is a meta-learning method that aims to learn a global initialization of parameters that encourage fast adaptation on new tasks. MAML uses episodic sampling of tasks at meta-training phase. At each task, the model parameters are initialized from the meta-learned parameters $\theta$ and are adapted to the task using one or few gradient steps. 
Even though MAML trains directly only on the support set during the meta-test phase (tasks sampled from $D_{\text{test}}$), it incorporates transductive batch normalization using the statistics of the query set. Therefore, for the purpose of grouping our methods, we categorize MAML as a transductive method. 

\textbf{LaplacianShot} LaplacianShot \cite{Laplacian} achieves label propagation on the query set by minimizing the sum of two terms. The first encourages query samples to bear the same label as the closest class prototype in the embedded space, where each prototype is computed as the centroid of support samples for each class. The second term is a pairwise Laplacian term that encourages nearby samples in the embedding space to be given similar assignments, hence effectively propagating labels. Note that contrary to most methods, LaplacianShot does not optimize for the class prototypes, but instead directly optimizes for the assignment variables of each query sample. The initial objective is optimized using bound optimization, which leads to a series of closed-form updates of the assignments, with convergence guarantees.  

\textbf{TIM} Authors in \cite{TIM} propose to use the query set by finding a classifier that maximizes the mutual information between the query features and the query predictions, while still properly classifying samples from the support set (i.e minimizing the cross-entropy between support ground truth labels and predictions of the classifier on the support set). Authors propose two schemes of optimization for their method, but we only reproduce the best performing version TIM-GD.

\section{Experiments}

\subsection{Evaluation Setup}
\label{sec:setup}
All methods are pre-trained for three seeds with a maximum of 100,000 iterations on the CRC-TP dataset. To ensure a fair comparison between methods, we performed all experiments of \autoref{tab:main_results} on the ResNet-18 model. We train the networks using Adam optimizer~\cite{kingma2014adam} with an initial learning rate set to $5\mathrm{e}{\text{-}4}$, and we use cosine decay. We apply random cropping, random flipping, and color jittering data augmentations. 
\MainTable
\textbf{Meta-Training:} For meta-learning methods (MAML\cite{maml}, Protonet\cite{prototypical_nets}, and MetaOpt\cite{lee2019meta}), we perform episodic training with 5-way 5-shot tasks in each iteration.

\textbf{Standard-Training:} We train a single model with cross-entropy supervision for all the methods using non-episodic training (SimpleShot\cite{simpleshot}, Distill\cite{tian2020rethinking}, Finetune\cite{closer_look}, TIM\cite{TIM}, LaplacianShot\cite{Laplacian}). Each iteration is performed on a batch of 100 samples. To perform self-distillation in the Distill\cite{tian2020rethinking} method, we train the network for another 100,000 iterations with the same learning schedule. For training, a machine with single NVIDIA V100 GPU with 16 GB memory was used.

During all the training procedures, we randomly sampled 250 tasks in a 5-way 5-shot setting from the NCT dataset to evaluate the method after every 1000 iterations (for the experiments on NCT (Table~\ref{tab:main_results} CRC-TP $\rightarrow$ NCT) methods are evaluated using LC25000 as the validation set). We perform meta-testing on the best model with the highest accuracy on these randomly generated NCT validation sets.
To simplify experiments, no hyperparameter tuning was made, and it will remain as future work.

\textbf{Meta-Testing: } We perform meta-testing on 1000 randomly sampled tasks from NCT, LC25000, and BreakHis datasets for each \neardomain~, \middledomain~, and \outdomain~ scenarios. In all experiments, we use standard 5-way tasks with 15 query samples per class(example of a single task is shown in \autoref{fig:taskexample} in the Appendix section). We evaluate methods on 1-shot, 5-shot, and 10-shot settings and report the average accuracy and 95\% confidence intervals.

\subsection{Benchmark Results}
We compare different state-of-the-art few-shot learning methods on the three scenarios(CRC-TP $\rightarrow$ NCT , CRC-TP $\rightarrow$ LC25000, CRC-TP $\rightarrow$ BreakHis). We train our models with base classes of the CRC-TP dataset and then apply meta-testing on the new classes from the NCT, LC25000, and the BreakHis datasets. Results are presented in \autoref{tab:main_results}. As introduced in appendix \ref{sec:datasets}, BreakHis dataset consists of four magnification rates for image patches. In the following experiments, We considered these magnifications as four super classes, each containing eight sub classes (tissues) of the BreakHis dataset (Table \ref{tab:scales} presents results of testing each magnification level separately). It is clear from table \ref{tab:main_results} that episodic training methods are out-performed by non-episodic methods in both inductive and transductive settings. Protonet performs significantly better in 1-shot and 5-shot settings compared to MetaOpt. This is in line with the results reported in \cite{guo2020broader} for cross-domain few-shot benchmark. The best performing inductive method in almost all the scenarios is Finetune. It is only outperformed by Distill in the 5-way 1-shot setting of the CRC-TP $\rightarrow$ LC25000 scenario. TIM achieves the best results for both inductive and transductive settings in all three scenarios. It results in at least 5\% improvement among other methods in the 5-way 1-shot setting of CRC-TP $\rightarrow$ NCT and CRC-TP $\rightarrow$ LC25000. Please see Table~\ref{tab:main_results_nct} in the Appendix for the results considering NCT as the base training source dataset.

As shown in Table~\ref{tab:main_results}, moving from \neardomain~to \middledomain~and \outdomain~scenarios performance drops drastically in all 5-way(1-shot/5-shot/10-shot) settings. 
In the near-domain setting (CRC-TP $\rightarrow$ NCT), the new classes are semantically related to base classes (same cancer) but correspond to different levels of granularity. This near-domain setting is highly relevant in histology practices due to the fine granularity of cancer-related tissue classification tasks and the variety of data-preparation techniques, which result in disparities in the labels across clinical sites. In this near-domain setting, our experiments in Table~\ref{tab:main_results} show that state-of-the-art few-shot methods in the 5-way 5-shot setting could achieve performances close to or higher than 80\%,  approaching supervised-learning upper baselines. This suggests that few-shot learning is highly relevant in this specific setting, and has the potential to make an impact in practice. In the middle-domain scenario (CRC-TP $\rightarrow$ LC25000), three of the five classes are closely related to the base classes (same cancer) while the other two classes are lung tissues(see Figure~\ref{fig:lcexample} and Table~\ref{tab:datasets}). Results of the best performing methods are close to 80\%. Since there are two different cancers in the LC25000 dataset, it is simpler for the method to classify and separate tissues of one cancer from the other (see Figure~\ref{fig:t-sne}). In the out-domain setting (CRC-TP $\rightarrow$ BreakHis, which corresponds to different cancer types), our findings are, to some extent, in line with \cite{guo2020broader}. However,  \cite{guo2020broader} considers extreme domain shifts with the source and target datasets coming from completely different domains (e.g. Natural images vs X-Ray images). 
The best performing method in our CRC-TP $\rightarrow$ BreakHis 5-way 5-shot setting results in 60\% accuracy (Table~\ref{tab:backbones}). We believe that, even in our out-domain setting, this 60\% performance is a good starting point to further encourage research in the cross-domain few-shot learning paradigm. 

\subsection{Comparison of different backbones}
We evaluate the effect of using different backbone architectures for the best performing transductive and inductive methods, TIM\cite{TIM} and Finetune\cite{closer_look}. Apart from Resnet-18 which is presented in Table~\ref{tab:main_results}, we consider two other backbones; Wide-ResNet (WRN-28-10)\cite{zagoruyko2016wide} and DeiT-Tiny\cite{touvron2020deit}. The results are presented in Table~\ref{tab:backbones}. Same as our previous experiments in Table~\ref{tab:main_results} and Table~\ref{tab:main_results_nct}, TIM performs better with both WRN and DeiT-Tiny compared to the Finetune method. DeiT-Tiny$^R$ denotes a network with random weights. We observe that pre-training the DeiT-Tiny backbone on ImageNet (denoted by DeiT-Tiny$^\ast$) results in improved performance. For CRC-TP $\rightarrow$ NCT and CRC-TP $\rightarrow$ LC25000 5-way 10-shot settings results are close to 90\% which means with only 10 support samples the methods can approach supervised learning results.( See  Table~\ref{tab:datasets} for supervised learning results on each dataset.)
\BackboneTableNew

\section{Conclusion and Discussion}
\label{sec:conclusion}
 We introduced FHIST, a diverse and realistic benchmark for few-shot classification of histological images. We devised three few-shot adaptation scenarios, with increasing levels of domain shifts between the source and target datasets: 1. \neardomain, 2. \middledomain~and 3. \outdomain. We evaluated the state-of-the-art and well-known methods in few-shot learning on our new benchmark, in both the
 inductive and transductive settings. The studied methods included both meta-learning and non-episodic approaches. Our experiments reveal that popular and convoluted meta-learning methods 
 under-perform simple non-episodic techniques, which are based on finetuning/regularization at inference. Our finding is in line with recent observations in the context of few-shot classification of natural images with domains shifts ~\cite{closer_look}. We hope that the introduced FHIST dataset, along with the evaluations of state-of-the-art methods, inspires further 
 research in this direction, and help in building more realistic evaluations and fairer comparisons of few-shot learning methods. 
 In a future work, we intend to investigate the degree to which different tissue types correlate with each other, within and across datasets. We believe that 
 leveraging the relationship between tissue types would result into better generalization for few-shot classification of histology images.

\section*{Acknowledgements}
The authors would like to thank Compute Canada and Calcul Québec for computational resources, and NSERC for funding this study in part. 

\bibliographystyle{plain}
\bibliography{biblio}

\clearpage
\appendix

\section{Appendix}

\subsection{Datasets used in our benchmark}

A list of available public histology datasets and challenges are presented in Table~\ref{tab:histologydatasets} and Table~\ref{tab:challenges}. The following four datasets are selected for our benchmark:

\label{sec:datasets}
\textbf{BreakHis} The Breast Cancer Histopathological Image Classification (BreakHis) dataset \cite{spanhol2015dataset} was created through a clinical study from January 2014 to December 2014 in collaboration with the P\&D Laboratory, Brazil. The dataset is composed of 7,909 microscopic images of benign and malignant breast tumors that are obtained from 82 anonymous patients using four magnification rates: 40X, 100X, 200X, and 400X. 24 of these patients have benign breast cancer and the rest have malignant breast cancer \cite{spanhol2016breast}. The dataset contains 2,480 benign and 5,429 malignant samples. Histologically benign is a term referring to a lesion that does not match any criteria of malignancy whereas a malignant tumor is a synonym for cancer \cite{chun2015case, kjellin2014differentially}. According to the way the tumoral cells look under the microscope, both benign and malignant breast tumors are sorted into subclasses by pathologists. Therefore, there are four distinct types of benign breast tumors in the dataset, namely adenosis (A), fibroadenoma (F), phyllodes tumor (PT), and tubular adenona (TA), as well as  four distinct types of malignant tumors, namely carcinoma (DC), lobular carcinoma (LC), mucinous carcinoma (MC), and papillary carcinoma (PC). The  images are provided in PNG format and have 700$\times$460 pixel resolution with three-channel RGB (8-bit depth in each channel) (Figure~\ref{fig:breakhisexample}).

\textbf{CRC-TP} The Colorectal Cancer Tissue Phenotyping (CRC-TP) dataset \cite{javed2020multiplex} includes 280K patches extracted from 20 whole-slide images (WSIs) of CRC stained with Hematoxylin \& Eosin (H\&E). For each slide, which is obtained from a different patient, region-level annotation of seven distinct tissue phenotypes are provided by expert pathologists (KB and KH). Based on these annotations, patches were extracted from WSIs, and based on majority of its content, each patch was assigned a unique label. The patches that have significant pixels from more than one phenotype are excluded from the dataset, which means that patch of a particular phenotype in the dataset is mostly composed of one tissue phenotype. As a result, the dataset includes 50K patches for each of the Stroma (St), Complex Stroma (CS), Tumor (Tu), and Smooth Muscle (SM) phenotypes. The Inflammatory (In), Benign (Be), and Debris (De) phenotypes consist of 20K, 30K ,and 30K patches respectively. Following the \cite{kather2016multi}, the patch size is fixed to 150$\times$150 pixels extracted at 20$\times$ magnification level and the patches are non-overlapping( Figure~\ref{fig:crctpexample}).

\textbf{LC25000} The Lung and Colon Histopathological Image (LC25000) dataset \cite{borkowski2019lung} consists of 25,000 color images of benign and cancerous lung and colon tissue images with five classes of 5,000 images each. The  images have $768\times768$ pixel resolution and are in jpeg file format. This dataset was created after applying data augmentation to 500 total images of colon tissue (250 benign colon tissue and 250 colon adenocarcinomas) and 750 total images of lung tissue (250 benign lung tissue, 250 lung adenocarcinomas, and 250 lung squamous cell carcinomas). The following augmentations were applied to the 1250 colon and lung tissue images: horizontal and vertical flips (0.5 probability) and left and right rotations (up to 25 degrees, 1.0 probability)( Figure~\ref{fig:lcexample}).

\textbf{NCT-CRC-HE-100K} The NCT-CRC-HE-100K dataset \cite{kather100_data} includes 100,000 non-overlapping image patches extracted from H\&E stained human cancer tissue slides from formalin-fixed paraffin-embedded (FFPE) samples from the NCT Biobank (National Center for Tumor Diseases, Heidelberg, Germany) and the UMM pathology archive (University Medical Center Mannheim, Mannheim, Germany). The dataset includes 9 tissue classes, namely Adipose (ADI), background (BACK), debris (DEB), lymphocytes (LYM), mucus (MUC), smooth muscle (MUS), normal colon mucosa (NORM), cancer-associated stroma (STR), and colorectal adenocarcinoma epithelium (TUM). All images are $224\times224$ pixels in size at $0.5$ microns per pixel, and they are color-normalized using Macenko's method \cite{macenko2009method} ( Figure~\ref{fig:nctexample}). 
\HistologyDatasets
\ChallengesTable

\subsection{Results of base training with NCT}
We repeat all the experiments of Table~\ref{tab:main_results} by changing the base dataset to NCT which is the second-largest dataset in the benchmark with 100,000 patches. We set the learning rate value to $5\mathrm{e}{\text{-}5}$. Other parameters are the same as the previous experiments. Results are shown in Table~\ref{tab:main_results_nct}. Same as the results in Table~\ref{tab:main_results}, Finetune and TIM are the best performing methods in the inductive and transductive settings respectively. TIM achieves better results throughout the whole experiments. Episodic training methods are mostly out-performed by non-episodic methods in both inductive and transductive settings. We see a huge performance drop in the NCT $\rightarrow$ CRC-TP scenario.  In general, Compared to the results in Table~\ref{tab:main_results}, considering CRC-TP as the base dataset results in better performance. This is because CRC-TP has ~2.8 times more training examples than the NCT dataset. Hence, it's better to choose CRC-TP over NCT as the base dataset in the source domain. 
\MainTableNCT

\subsection{Effect of image scale}
As mentioned in Table~\ref{tab:datasets}, BreakHis contains images with different magnification factors(see Figure~\ref{fig:breakhisexample}). Results presented in Table~\ref{tab:main_results} and Table~\ref{tab:backbones} do not correct for various magnification factors. Here we conduct an experiment to study this effect. The results are presented in Table~\ref{tab:scales}. We consider 4 scales as these are the 4 possible scales in BreakHis. We observe that unifying magnification across BreakHis impedes the performance. 
\ScalesTable

\CRCFig
\LCFig
\NCTFig
\BHFig

\TaskFig






\end{document}